%% file: 0_cvpr24w_ID30.tex
\documentclass[10pt,twocolumn,letterpaper]{article}

\usepackage{cvpr}              

\input{preamble}

\definecolor{cvprblue}{rgb}{0.21,0.49,0.74}
\usepackage[pagebackref,breaklinks,colorlinks,citecolor=cvprblue]{hyperref}

\usepackage[capitalize]{cleveref}
\crefname{section}{Sec.}{Secs.}
\Crefname{section}{Section}{Sections}
\Crefname{table}{Table}{Tables}
\crefname{table}{Tab.}{Tabs.}

\usepackage{pifont}
\usepackage{bbm}
\usepackage{caption}
\usepackage{subcaption}
\usepackage{enumitem} 
\usepackage{xspace}
\usepackage{makecell}
\usepackage{graphicx}
\usepackage{booktabs}
\usepackage{multirow}
\usepackage{tikz}

\def\ie{\emph{i.e.}}

\newcommand{\oursbig}{Unsupervised Video Instance Segmentation (UVIS)\xspace}
\newcommand{\oursshort}{UVIS\xspace}

\newcommand{\cmark}{\ding{51}}%
\newcommand{\xmark}{\ding{55}}%
\newcommand{\more}[1]{\small{\textcolor[RGB]{57,181,74}{#1}}}

\def\paperID{30} 
\def\confName{CVPR}
\def\confYear{2024}

\title{UVIS: Unsupervised Video Instance Segmentation}

\author{Shuaiyi Huang$^{1}$, Saksham Suri$^{1}$, Kamal Gupta$^{1}$\thanks{Work done while at UMD.}~, \\Sai Saketh Rambhatla$^{2}$\protect\footnotemark[1]~, Ser-nam Lim$^{3}$, Abhinav Shrivastava$^{1}$\\
$^{1}$University of Maryland, College Park \quad $^{2}$Meta \quad $^{3}$University of Central Florida
}

\begin{document}

\maketitle

\begin{abstract}
Video instance segmentation requires classifying, segmenting, and tracking every object across video frames. Unlike existing approaches that rely on masks, boxes, or category labels, we propose \oursshort, a novel \oursbig framework that can perform video instance segmentation without any video annotations or dense label-based pretraining. Our key insight comes from leveraging the dense shape prior from the self-supervised vision foundation model DINO and the open-set recognition ability from the image-caption supervised vision-language model CLIP. Our \oursshort framework consists of three essential steps: frame-level pseudo-label generation, transformer-based VIS model training, and query-based tracking. To improve the quality of VIS predictions in the unsupervised setup, we introduce a dual-memory design. This design includes a semantic memory bank for generating accurate pseudo-labels and a tracking memory bank for maintaining temporal consistency in object tracks. We evaluate our approach on three standard VIS benchmarks, namely YoutubeVIS-2019, YoutubeVIS-2021, and Occluded VIS. Our \oursshort achieves 21.1 AP on YoutubeVIS-2019 without any video annotations or dense pretraining, demonstrating the potential of our unsupervised VIS framework.
\end{abstract}

\input{./1_introduction}

\input{./2_related}

\input{./3_method}

\input{./4_experiments}

\input{./5_conclusion}

{
    \small
    \bibliographystyle{ieeenat_fullname}
    \bibliography{references}
}


\end{document}


\maketitle

In this supplementary material, we provide more detailed quantitative results and qualitative analysis of our method as follows: i) In Sec.~\hyperref[sec:supp_A]{A}, we present F1-score statistics on the train set to assess the quality of pseudo-labels, in addition to the prototype memory filtering (PMF) ablation discussed in the main paper. ii) In Sec.~\hyperref[sec:supp_B]{B}, we offer more insights into our implementation by providing details on class-agnostic mask generation, the prompts used for text-instance matching, and additional experimental details that complement the information provided in the main paper. iii) Sec.~\hyperref[sec:supp_C]{C} showcases more qualitative results on Youtube-VIS 2019~\cite{yang2019video}, Youtube-VIS 2021~\cite{yang2019video} and OVIS~\cite{qi2021occluded} validation set. For more qualitative video results, please refer to our uploaded video file in the supplementary material.

\section*{A. PMF Impact on Pseudo-Label Quality}
\label{sec:supp_A}
\vspace{-2mm}
To evaluate the impact of our PMF on pseudo-label quality, we conducted an analysis by computing the F1 score between the filtered pseudo-labels and the per-frame ground truth on the Youtube-VIS 2019 train set~\cite{yang2019video}. The F1 score provides insights into the removal of false positives while maintaining true positives. For evaluation, we consider a prediction as a true positive if its mask IoU with the corresponding ground truth mask is above 0.5. Table~\ref{tab:supp_f1} summarizes the results, including per-class F1 scores and the averaged F1 score over all categories (mF1).  Comparing the results to the case without filtering (28.6\%), the score-based filtering (mask and CLIP score with a 0.7 threshold) improves the mF1 to 42.5\%. With the integration of our PMF, we achieve a further improvement to 43.1\%, obtaining the highest F1 score across the majority of classes. These findings indicate that our PMF enhances the quality of pseudo-labels, demonstrating its effectiveness in improving VIS results.

\begin{table*}[h]
	\centering
	\footnotesize
	\renewcommand{\arraystretch}{1.1}
    \renewcommand{\tabcolsep}{2pt}
	
	\resizebox{1.0\textwidth}{!}{
  \begin{tabular}{lccccccccccccccccccccc}
			\toprule
			 Filtering Methods  & person & panda &lizard & parrot &skateboard& sedan & ape &dog& snake&monkey&hand&rabbit&duck&cat&cow&fish&train&horse&turtle&bear&\textbf{mF1}(\%)  \\  
            None &\textbf{18.3} & 36.4 & 41.1 & 43.0 & \textbf{1.7} & 28.1 & 34.2 & 20.5 & 14.3 & 36.4 & 8.8 & 39.4 &39.0 & 22.3 & 31.6 & 23.9 & 20.7 & 30.4 & 46.8 & 28.7 & -\\
            
		score-based &8.2 & 47.5 & 56.3 & 58.5 & 1.6 & \textbf{35.8} & 47.0 & 37.2 & 35.3 & 46.4 & 9.9 & 53.8 &\textbf{52.3} & 50.2 & 54.0 & 42.0 & 43.3 & 52.5 & \textbf{63.9} & 56.3 & -\\

  	score-based + PMF &5.9 & \textbf{47.9} & \textbf{57.6} & \textbf{59.8} & 1.6 & 34.7 & \textbf{47.2} & \textbf{40.0} & \textbf{39.7} & \textbf{47.0} & \textbf{10.0} & \textbf{53.8} &52.0 & \textbf{51.9} & \textbf{55.6} & \textbf{43.1} & \textbf{45.7} & \textbf{53.2} & 63.5 & \textbf{57.3} & -\\

  \midrule
  	Filtering Methods  & motorbike &giraffe & leopard &fox & deer & owl & surfboard & airplane & truck & zebra & tiger & elephant & snowboard & boat & shark & mouse & frog & eagle & seal & tennis racket &\textbf{mF1}(\%)  \\  
        None &18.7 & 43.2 & 44.0 & 44.6 & 20.4 & 51.0 & 1.5 & 22.6 & 30.0 & 35.2 & 44.4 & 43.9 &\textbf{0.2} & 20.0 & 22.7 & 26.0 & 37.5 & 33.4 & 31.5 & 6.4 & 28.6\\
            
		score-based &35.4 & \textbf{59.8} & \textbf{52.0} & \textbf{55.7} & \textbf{15.0} & \textbf{65.9} & \textbf{2.5} & 46.5 & 48.8 & 58.2 & \textbf{57.7} & 63.0 &0.1 & 29.2 & 40.8 & 50.1 & \textbf{51.0} & 58.2 & 49.7 & 8.5 & 42.5\\

  score-based + PMF & \textbf{36.9} & 59.6 & 51.4 & 55.2 & 14.7 & 65.4 & 2.4 & \textbf{48.9} & \textbf{50.0} & \textbf{60.4} & 57.2 & \textbf{64.9} & 0.1 & \textbf{30.3} & \textbf{41.4} & \textbf{50.2} & 49.6 & \textbf{58.3} & \textbf{49.7} & \textbf{9.0} & \textbf{43.1}\\
			\bottomrule
	\end{tabular}}
     \caption{\textbf{Per-class and overall F1-score results for pseudo-labels filtering on the Youtube-VIS 2019 train set~\cite{yang2019video}.} F1-scores are obtained using three different filtering methods: without any filtering (row 1), filtering by mask and clip threshold (row 2), and our prototype memory filtering (PMF) method (row 3). The best-performing results are highlighted in bold. Among the methods, our PMF approach achieves the highest mean F1-score across multiple classes, indicating its effectiveness in reducing false positives while preserving true positives.}
	\label{tab:supp_f1}
\end{table*}

\vspace{-2mm}
\section*{B. More Implementation Details}
\label{sec:supp_B}
\vspace{-3mm}
\noindent\textbf{B1. Class-agnostic mask generation}
Our approach requires pseudo-labels that include both regions of interest and their corresponding labels. To generate possible object regions and their corresponding masks we use an existing off-the-shelf unsupervised approach called CutLER~\cite{wang2023cut}. CutLER is trained in a unsupervised manner using coarse masks obtained from the self-supervised DINO~\cite{caron2021emerging} model for the ImageNet~\cite{deng2009imagenet} dataset. These masks are then used to train a Cascade Mask R-CNN~\cite{cai2019cascade} backbone in a class agnostic manner. The trained detector referred to as CutLER shows good generalization in predicting masks and boxes around objects in our work. For each frame $V_{t}$, CutLER predicts a set of boxes $\{b^i_{t}\}$, masks $\{M^i_{t}\}$ and corresponding objectness scores $\{o^i_{t}\}$ where $i$ corresponds to the $i^\text{th}$ object instance in the frame. We use a threshold of $0.7$ to filter out low confidence predictions for this step. More details about the training and generalizations of CutLER can be found in their paper~\cite{wang2023cut}.


\noindent\textbf{B2. Prompts for CLIP-based Text-Instance Matching}
In CLIP-based Text-Instance Matching, an instance crop image is assigned a class label by computing the cosine similarity between the image embedding and a set of text prompts. The text prompts are generated given the dataset label set using simple string templates such as ``a photo of $<$ class $>$". Multiple prompts per class are typically used to increase coverage. Specifically, the template ``a photo of $<$ class $>$", along with the following six variations ``a photo of $<$ class $>$ doing", ``a photo of $<$ class $>$ moving", ``a photo of $<$ class $>$ with", ``a photo of $<$ class $>$ on", ``a photo of $<$ class $>$ in", and ``a photo of $<$ class $>$ at" are employed for each class. The model selects the closest matching prompt based on cosine similarity, thereby assigning the corresponding class label to the instance crop image.

\noindent\textbf{B3. Architecture and Optimization}
In our implementation, we adopt Detectron2~\cite{wu2019detectron2} and adhere to the settings proposed in MinVIS~\cite{huang2022minvis} for video instance segmentation. Our chosen architecture consists of six multi-scale deformable attention Transformer (MSDeformAttn)~\cite{zhu2020deformable} layers applied to feature maps at resolutions 1/8, 1/16, and 1/32. Additionally, we incorporate a simple upsampling layer with lateral connection to generate the final 1/4 resolution feature map, which serves as the per-pixel embedding. For the transformer decoder, we employ 9 layers and set the number of queries to 100 by default. During optimization, we assign a weight of 2.0 to the classification loss ($\mathcal{L}_\text{cls}$) and 5.0 to the segmentation loss ($\mathcal{L}_\text{seg}$). We utilize the AdamW optimizer with an initial learning rate of 0.0001 and employ a step learning rate schedule. In our unsupervised setup, we keep the backbone fixed. During inference, we retain the top 10 predictions for each video sequence.

\section*{C. More Qualitative Results}
\label{sec:supp_C}
\begin{figure*}[b]
	\centering
	\includegraphics[width=\linewidth]{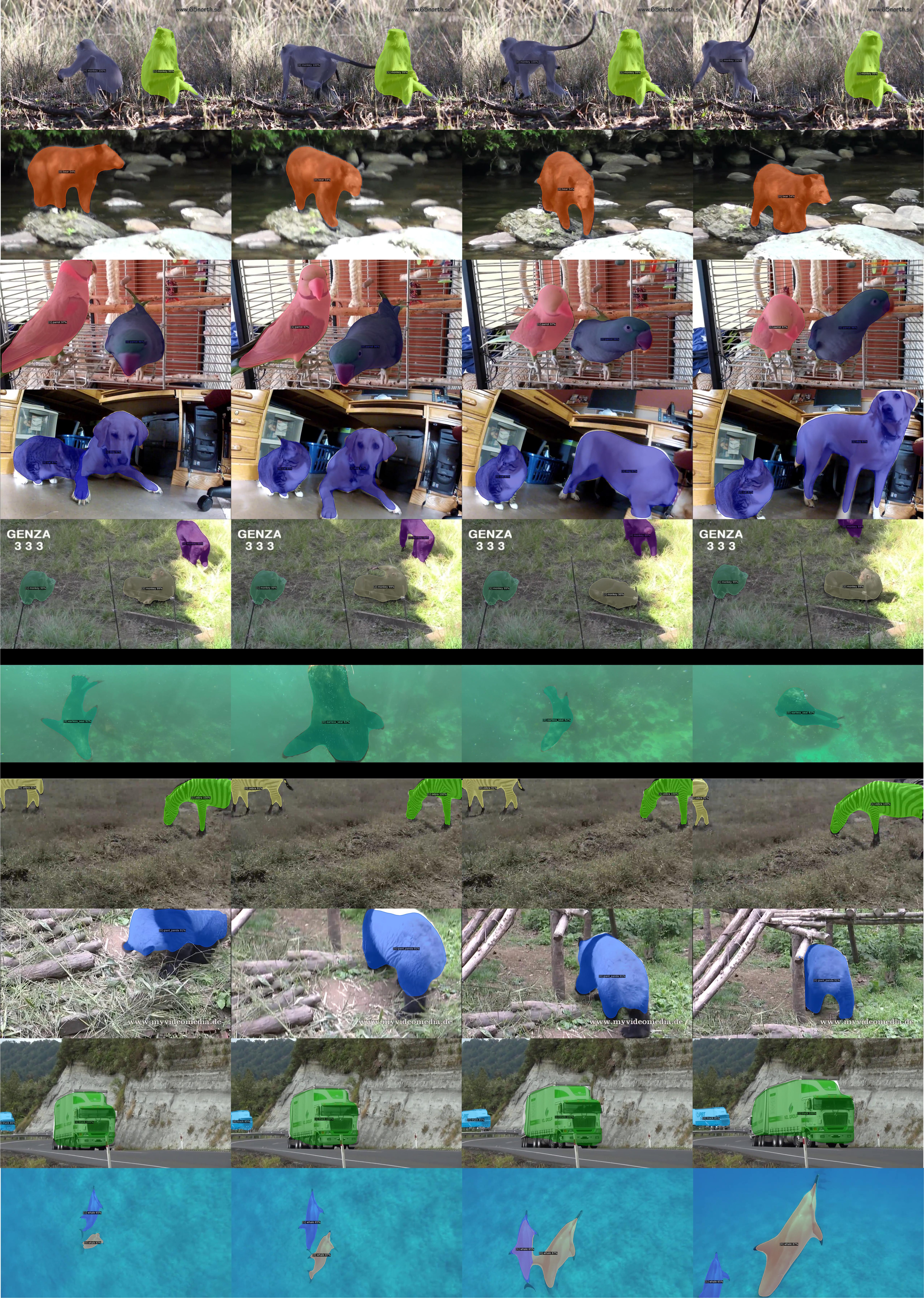}
	\caption{\small\textbf{Qualitative results of our UVIS on Youtube-VIS 2021~\cite{yang2019video} validation set.} }
	\label{fig:supp_vis_ytvis21}
\end{figure*}

\begin{figure*}[b]
	\centering
	\includegraphics[width=\linewidth]{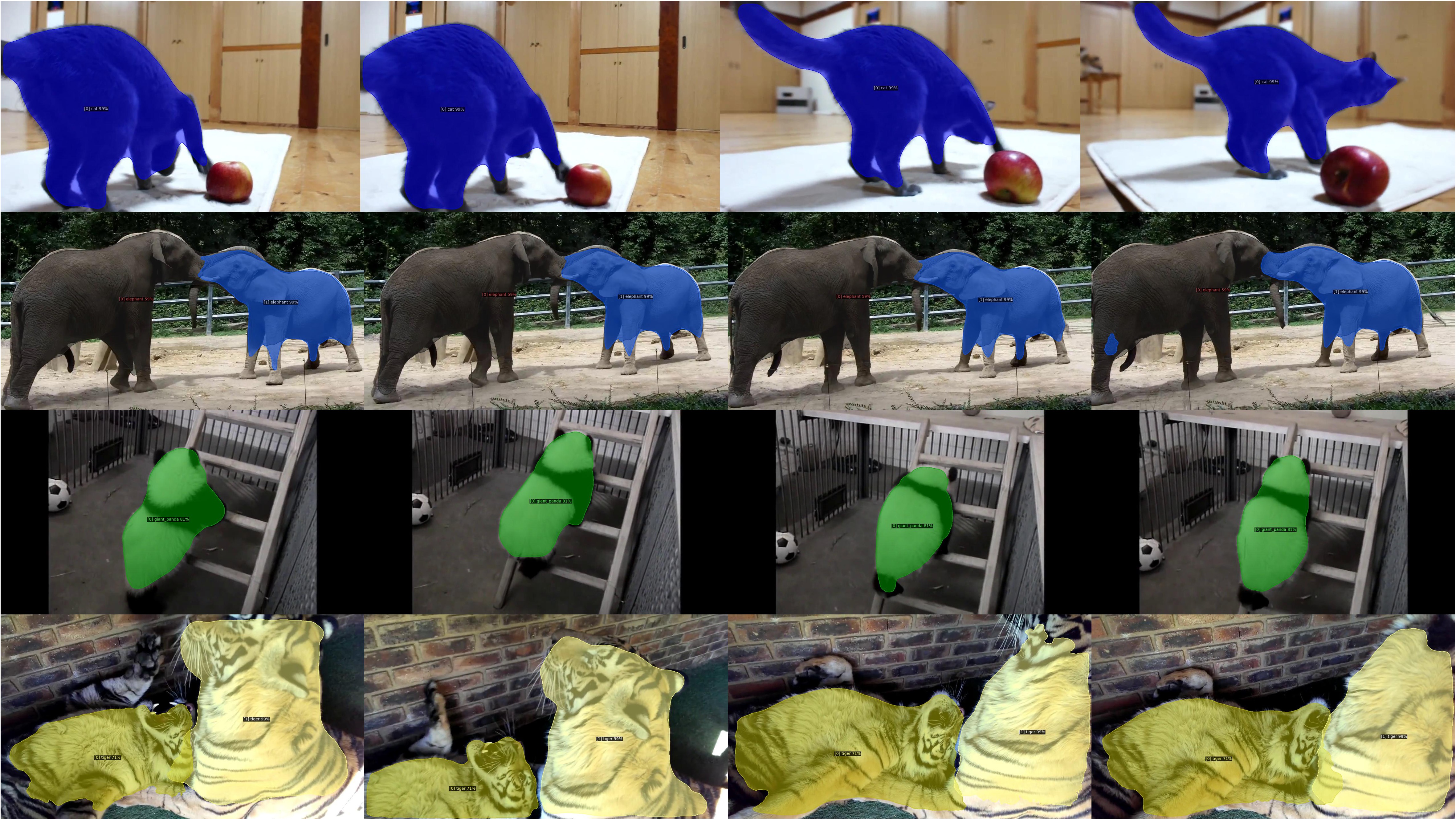}
	\caption{\small\textbf{Qualitative results of our UVIS on Youtube-VIS 2019~\cite{yang2019video} validation set.} }
	\label{fig:supp_vis_ytvis19}
\end{figure*}

\begin{figure*}[b]
	\centering
	\includegraphics[width=\linewidth]{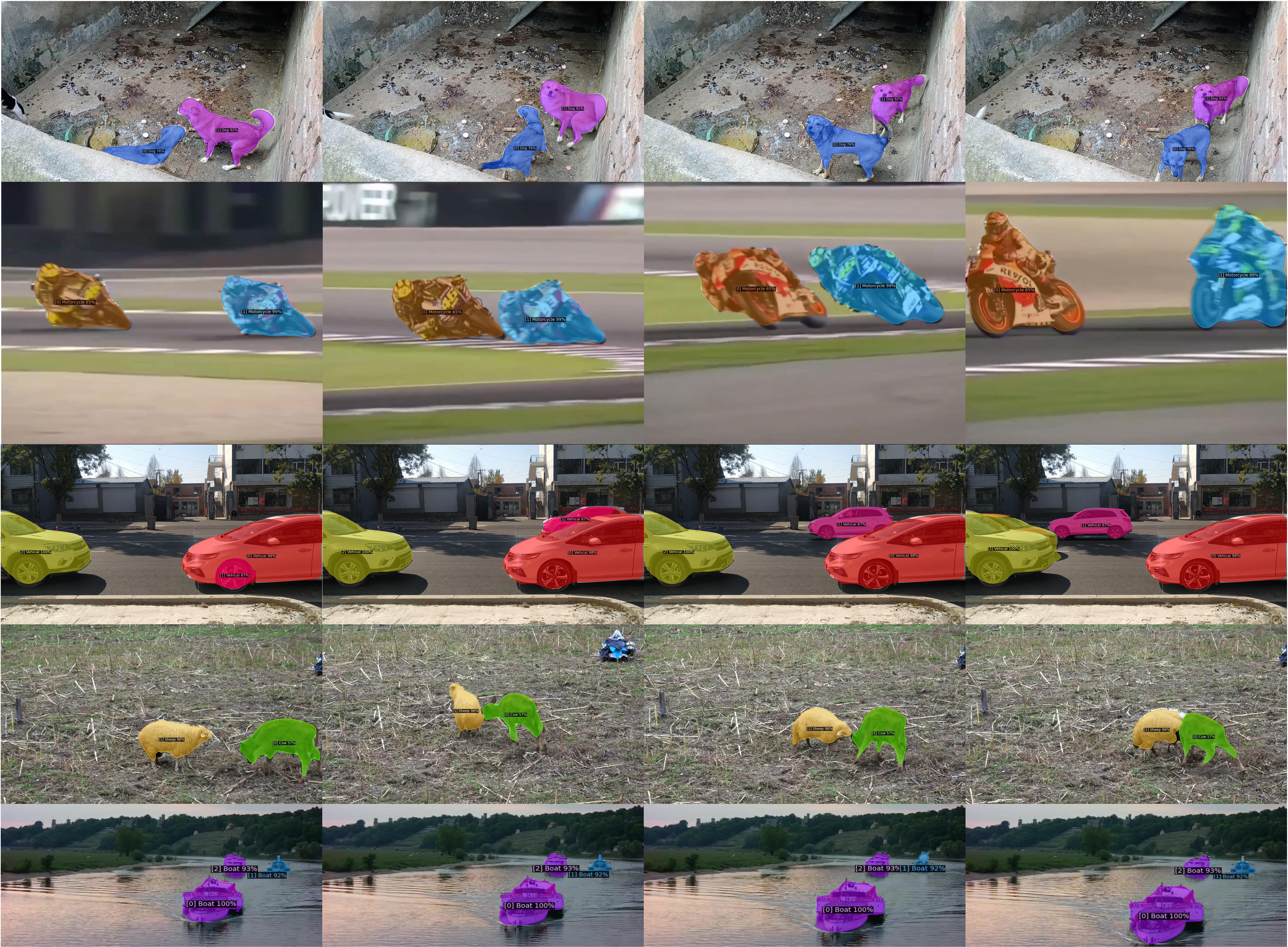}
	\caption{\small\textbf{Qualitative results of our UVIS on OVIS~\cite{qi2021occluded} validation set.} }
	\label{fig:supp_vis_ovis}
\end{figure*}

\vspace{-3mm}
More qualitative results from the predictions of our UVIS on Youtube-VIS 2019~\cite{yang2019video}, Youtube-VIS 2021~\cite{yang2019video} and OVIS~\cite{qi2021occluded} validation set, are shown in Figure~\ref{fig:supp_vis_ytvis19}, ~\ref{fig:supp_vis_ytvis21} and ~\ref{fig:supp_vis_ovis}, respectively. For more qualitative video results, please refer to our uploaded video file in the supplementary material.

{
    \small
    \bibliographystyle{ieeenat_fullname}
    \bibliography{references}
}


%% file: preamble.tex
\usepackage[dvipsnames]{xcolor}


%% file: 1_introduction.tex
\section{Introduction}
\label{sec:introduction}

\begin{figure*}[t]
  \centering
   \includegraphics[width=1.0\linewidth]{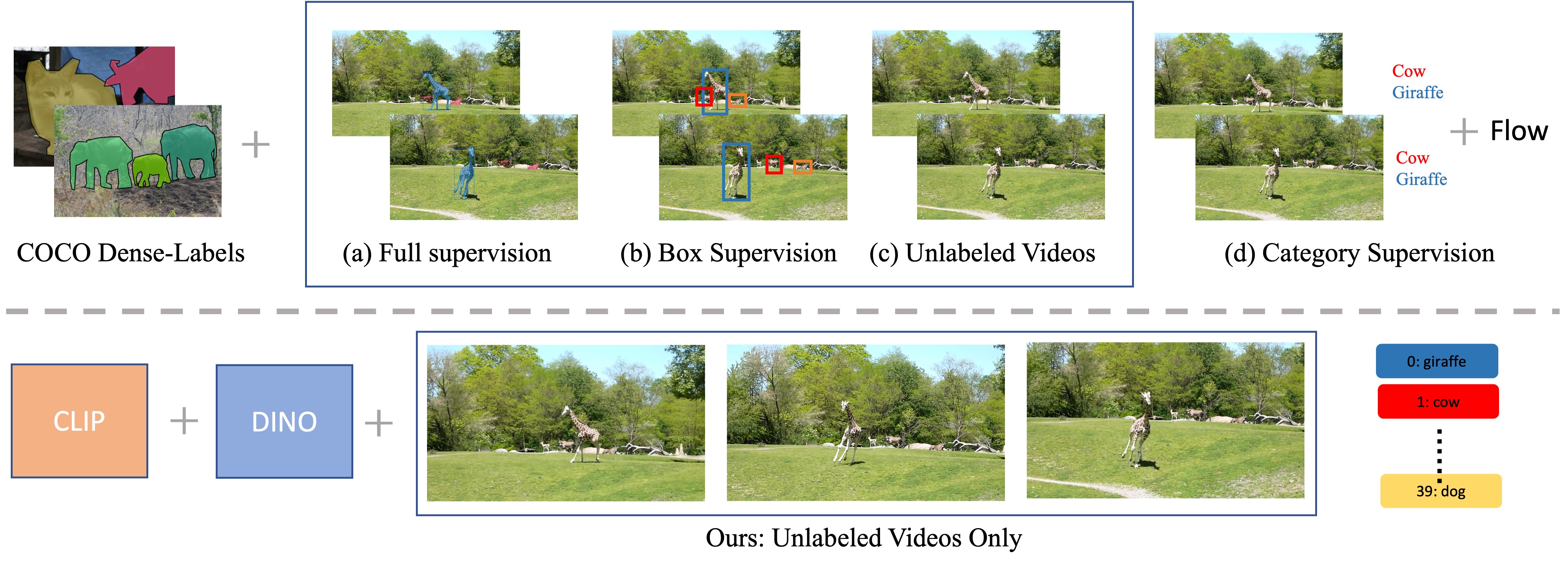}
    \caption{\textbf{Setting Overview.} Previous approaches have tried to use COCO dense annotations in addition to VIS dataset full supervision (a), box supervision (b) and no supervision (c). Additionally, previous works have also used flow information along with frame-level category labels (d). Our approach UVIS works in the unsupervised setting and does not require any dense labels or per frame labels and instead utilizes foundation models.}
  \label{fig:teaser}
\end{figure*}

Video Instance Segmentation (VIS)~\cite{yang2019video} is the task of classifying, segmenting, and tracking individual objects within a video, with a wide range of industry applications such as in robotics, sports, autonomous driving, surveillance, AR/VR, 3D navigation~\cite{shah2020ving,wei2023imitation,zheng2024texttt,zheng2024premier,zheng2023is,wang2023coplanner,xu2024drm,zheng2024prise,ji2024ace,sun2022transfer,zhu2021deep}, etc. It is a challenging problem due to variations in object appearances, occlusions, and cluttered scenes over time. Reliable models for VIS require dense annotated data which is costly. To circumvent the need for costly dense annotations in videos, existing methods have utilized various strategies such as pretraining on densely-labeled image datasets like COCO and finetuning on fully-labeled~\cite{cheng2021mask2former} or unlabeled~\cite{fu2021learning} videos, or reducing annotations through subsampled frames~\cite{huang2022minvis}, boxes~\cite{ke2023mask}, per-frame category labels~\cite{liu2021weakly}. However, these methods still rely on annotations~\cite{huang2022minvis,ke2023mask,liu2021weakly} or can only handle categories that overlap with the densely-labeled image dataset~\cite{fu2021learning}. In contrast, the human perception leverages image and video-level priors to effortlessly recognize, segment, and track objects~\cite{biederman1987recognition}. This leads us to explore whether it is possible to learn an unsupervised video instance segmentation model without any dense pretraining or video annotations, covering all categories in a dataset.

Unsupervised video instance segmentation presents several challenges when only the category label set is provided for the video dataset. The first challenge is accurately predicting object boundaries without dense labeling in videos. The second challenge is conducting object classification when only the category label set is available. To address these challenges, we draw inspiration from recent advancements in large-scale unsupervised vision models, specifically the dense shape prior in self-supervised vision model DINO~\cite{caron2021emerging} and the open-set recognition capability in image-caption supervised vision and language model CLIP~\cite{Radford2021LearningTV}. By combining these strengths, our unsupervised VIS model can effectively segment and recognize objects within a given vocabulary set without the need for dense pretraining. To the best of our knowledge, we are the first work to explore CLIP and DINO in the field of VIS. This naturally solves the limitation of existing works that can only handle categories that overlap with densely labeled external image datasets~\cite{fu2021learning}.

To this end, we introduce an unsupervised video instance segmentation framework (UVIS), which is the first VIS framework, that can learn to segment all categories in videos without any dense annotation based pretraining or video annotations, as shown in Figure~\ref{fig:teaser}. Our unsupervised framework for video instance segmentation comprises of three essential steps. First, we generate class-agnostic instance masks for each video frame using a pre-trained self-supervised model~\cite{wang2023cut} and equip the masks with semantic labels by using CLIP~\cite{radford2021learning}. Second, we train a transformer-based video instance segmentation model by using the per-frame pseudo-masks obtained from the first step~\cite{huang2022minvis}. Third, during inference, we generate dense and consistent mask tubes by linking per-frame predictions using bipartite matching of query features.

To generate high-quality VIS predictions in the unsupervised setup, we further propose a novel dual-memory design on top of the above proposed UVIS framework for semantically accurate and temporally-consistent predictions. Specifically, to obtain semantically accurate pseudo-labels, we construct a class-specific prototype memory bank during pseudo-label generation. The prototypes serve as representative references, enhancing generalization and handling noisy false positives. In addition, to address the inherent limitations of the online inference pipeline in VIS that utilized only short-term information for tracking, we propose a simple but effective tracking memory that models long-term temporal information. To summarize, our main contributions are as follows:

\begin{itemize}[leftmargin=*,itemsep=0in]
    \item We introduce the first VIS framework that eliminates the need for any video annotations or dense label based pretraining, thereby significantly reducing annotation costs. Our framework covers all categories in the dataset, offering a comprehensive solution.
    \item We propose a novel dual-memory design on top of our unsupervised VIS framework. This design includes a prototype memory filtering component, which enhances the quality of pseudo-labels, and a tracking memory bank, which captures long-term temporal information for accurate tracking.
    \item We conduct comprehensive experiments in three standard video instance segmentation datasets including YoutubeVIS-2019~\cite{yang2019video}, YoutubeVIS-2021~\cite{yang2019video}, and Occluded VIS~\cite{qi2021occluded}, demonstrating the potential of our unsupervised VIS framework.
\end{itemize}

%% file: 2_related.tex
\vspace{-0.1in}
\section{Related Work}
\label{sec:related}
\vspace{-0.05in}

\noindent\textbf{Video object segmentation.} Video object segmentation (VOS)~\cite{Perazzi2016, Pont-Tuset_arXiv_2017, caelles20182018, caelles20192019, xu2018youtube} is a dense binary classification problem of separating salient foreground objects from the background in videos. The most popular task in VOS is the so-called \textit{Semi-supervised} VOS, where the goal is to segment objects in target frames given ground truth masks in the first frame. To prevent the annotation costs of exhaustively labeling each frame, several weakly and unsupervised VOS method have been proposed. \cite{10.1145/3506716} uses video level tags while~\cite{8650146} uses point supervision as weak labels to train a weakly supervised VOS system. \cite{Zhen2020LearningDF,Lee2022UnsupervisedVO, TMO, ren2020rtnet} propose unsupervised VOS to completely eliminate the need for supervision and is a much harder problem than fully and weakly supervised VOS.
In this work, we tackle a much harder problem of unsupervised Video Instance segmentation that not only does background separation but additionally performs instance segmentation that requires classification and tracking without any human supervision.

\noindent\textbf{Supervised video instance segmentation.} Video Instance Segmentation (VIS), initially proposed by Yang et al.\cite{yang2019video}, is an extension of image instance segmentation to videos, where the goal is to classify, segment and track objects across video frames. Early approaches\cite{fu2020compfeat,maag2021improving,yang2021crossover,zhou2021target,li2021spatial} segment and classify objects in each frame independently, and then associated the objects across frames using heuristics such as mask or box IoU. Recently, transformer-based approaches for VIS have gained significant attention~\cite{hwang2021video,wang2021end,yang2021tracking,cheng2021mask2former}. These approaches train VIS models in a video-based manner where they feed a clip as input and generate spatio-temporal mask predictions in one shot. A more recent development is the introduction of MinVIS~\cite{huang2022minvis}. This pioneering work demonstrates that a transformer-based VIS model trained solely on images can achieve competitive performance without video-based training or specialized video-based architecture design. They observe that instance tracking naturally emerges in query-based image instance segmentation models with proper architectural constraints. We build our work on top of MinVIS~\cite{huang2022minvis} due to its excellent performance in VIS using image-based training. Note that such a pipeline differs fundamentally from existing approaches such as IDOL~\cite{IDOL}, which rely on post-processing steps like non-maximum suppression (NMS) during inference for tracking. However, MinVIS does not consider long-term temporal information during tracking, we address this inherent limitations by incorporating crucial temporal information in image-based VIS.


\noindent\textbf{Weakly/Semi-supervised video segmentation.} Reducing the annotation requirements in VIS has become a focus of recent research efforts~\cite{liu2021weakly,fu2021learning,huang2022minvis}. Liu \textit{et al.}\cite{liu2021weakly} utilize per-frame category annotations and correspondences~\cite{ahn2019weakly,he2023towards,huang2019dynamic,huang2022learning} in videos, but exhibiting limited competitiveness compared to supervised approaches. Fu~\textit{et al.}\cite{fu2021learning} utilize instance segmentation annotations from the COCO dataset to learn VIS without video annotations, but are only applicable to overlapping categories between video and image datasets. Huang~\textit{el al.}~\cite{huang2022minvis} utilized annotations in sub-sampled frames but still rely on dense annotations. Huang~\textit{el al.}~\cite{huang2024point} utilized point supervision in videos. In contrast, our \oursshort method handles all categories for a given vocabulary without any per-frame category/box/mask label or COCO pretraining. To the best of our knowledge, this is the first unsupervised VIS framework that achieves impressive results without any human annotations.

\noindent\textbf{VL models based segmentation.} Recently, foundational models trained on large amounts of uni-modal or multi-modal data using weak or self-supervision have gained significant attention~\cite{Bommasani2021FoundationModels}. CLIP~\cite{Radford2021LearningTV}, a vision-language model using image-text pairs as supervision, has been particularly popular. CLIP has been extended to perform per-pixel detection and segmentation tasks in images\cite{lueddecke22_cvpr,zhou2022zegclip,Luo2022SegCLIPPA,Zhou2021ExtractFD}. However, the effectiveness of CLIP for videos and instance segmentation tasks has not been thoroughly studied. In this work, we explore the use of CLIP for unsupervised VIS, which has not been adequately explored. DINO~\cite{caron2021emerging}, a uni-modal foundational model trained on unlabeled images using self-supervised learning, demonstrates impressive segmentation capabilities. However, it cannot handle complex tasks like instance segmentation due to the lack of labeled information. Our approach combines the segmentation capabilities of self-supervised models with the zero-shot capabilities of CLIP to perform instance segmentation in videos. While NamedMask~\cite{shin2022namedmask} is a related approach that performs semantic segmentation on images, our approach specifically focuses on VIS, offering a more comprehensive solution.

%% file: 3_method.tex
\vspace{-0.05in}
\section{Method}
\label{sec:method}

\begin{figure*}[t]
  \centering
   \includegraphics[width=1.0\linewidth]{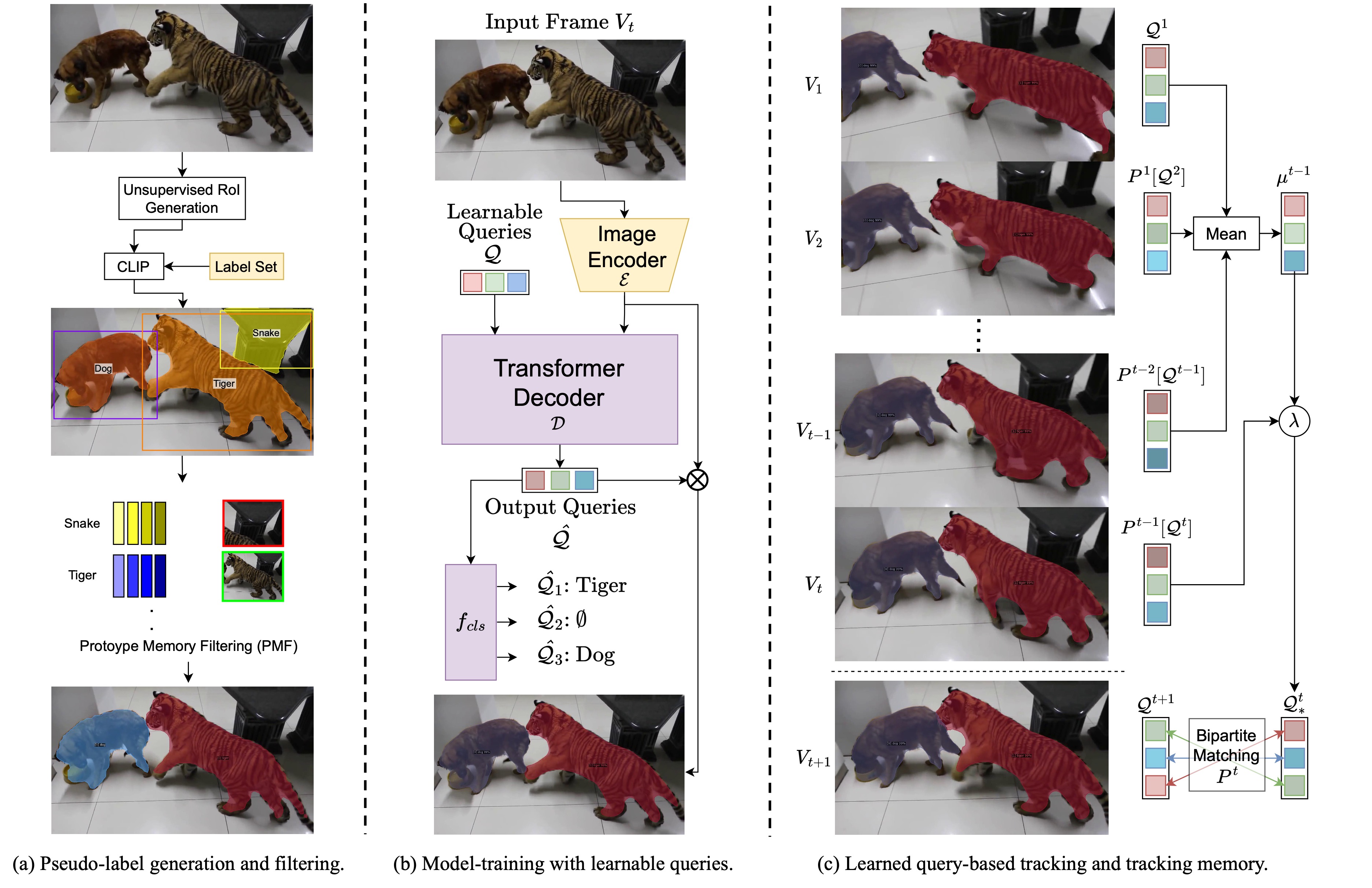}
     \footnotesize
  \caption{\textbf{We present our approach UVIS.} On the left we show our pseudo-label generation pipeline which involves generating masks and instance labels using CutLER~\cite{wang2023cut} and CLIP~\cite{radford2021learning} followed by Prototype Memory Filtering (PMF). In the center we show our model training which uses and image encoder and a transformer decoder to learn queries to predict per-frame predictions. On the right we show our proposed tracking memory approach which utilizes per frame queries and a memory based update rule to perform matching between frames to track instances and generate temporally consistent predictions.}
  \label{fig:combined}
\end{figure*}

Our objective is to learn a video instance segmentation model without groundtruth mask, box, or point annotations.
The problem is challenging since we need to maintain temporal consistency while the objects may undergo appearance changes, occlusions, or partial visibility, making it difficult to track and segment them accurately over time.
We build upon the recent advances in large-scale models pre-trained with Internet-scale data without any dense labels, also often called `foundation' models.
Many of these models are image and text-based and do not extend trivially to videos.
Hence, in this section, we propose the framework to utilize them for the video segmentation task.
Our framework consists of three steps as shown in Figure~\ref{fig:combined}:
(1) we start by generating pseudo-masks (\Cref{subsec:pseudo}) per video frame and build a prototype memory bank for different classes in the training data. Our proposed prototype memory encodes per-class semantic information and is used to filter the false positives improving the quality of the pseudo-labels, as shown in Figure~\ref{fig:combined}~(a);
(2) Secondly, we train a transformer-based video instance segmentation model (\Cref{subsec:mt}) by using the per-frame pseudo-masks generated from the first step as shown in Figure~\ref{fig:combined}~(b);
(3) Finally, during inference, we perform bipartite matching between instances of consecutive frames and propose a tracking memory (Section.~\ref{subsec:tracking}) to build dense and consistent mask tubes across the video as shown in Figure~\ref{fig:combined}~(c).

Formally, we are given a collection of $N$ videos $\mathcal{V} = \{V_n\}_{n=1}^{N}$ (with no pixel, box, or instance-level annotations), and a set of categories that we want to segment in these videos as $\mathcal{C} = \{l_c\}_{c=1}^{C}$, where $C=|\mathcal{C}|$ is the number of categories, and $l$ is the text label.
Note that we assume that no per-video label information is provided, \ie, the set of videos are not tagged with labels. Furthermore, a video may have zero or more of object instances corresponding to each label.
Each video $V \in \mathcal{V}$ can also have a variable number of frames and we denote the $t^\text{th}$ frame for this video as $V_{t}$. 


\subsection{Generating pseudo-labels for instance masks}
\label{subsec:pseudo}

\noindent\textbf{Class agnostic mask generation.} In the first step of our approach, we leverage self-supervised image models to generate pseudo-labels for video frames.
Self-supervised learning (SSL) models such as~\cite{caron2021emerging,he2022masked,chen2021empirical,chen2020big,chen2021exploring,zbontar2021barlow}, are typically trained using unlabeled ImageNet~\cite{deng2009imagenet} train set, and have an innate discriminative and localization abilities.
Several methods~\cite{wang2022tokencut,amir2021deep,LOST,rambhatla2023most} have been proposed to extract object masks from images using features from the SSL models. These methods typically work by performing a graph partitioning over features corresponding to various images patches and iteratively refining these partitions.
We adopt CutLER~\cite{wang2023cut}'s self-training strategy to generate mask and box predictions for each image using the pre-trained SSL model DINO~\cite{caron2021emerging}. See the supplementary material for the details of the approach.
Given a video frame $V_{t}$, CutLER predicts a set of boxes $\{b^i_{t}\}$, masks $\{M^i_{t}\}$ and their corresponding objectness scores $\{o^i_{t}\}$ where $i$ corresponds to the $i^\text{th}$ object instance in the frame.

\noindent\textbf{CLIP based Text-Instance Matching.}
In order to associate each mask $M^i_{t}$ to the corresponding label of interest, we utilize CLIP~\cite{radford2021learning}, a vision-language model trained with aligned text and image data. 
CLIP consists of a vision module $f^\text{CLIP}_\text{vision}$ and a text module $f^\text{CLIP}_\text{text}$ to compute image and text embeddings respectively.
Given an image $I$, the model assigns a class (from a list of classes) to it by computing the cosine similarity between the image embedding and the embeddings of a list of text prompts, and selecting the closest prompt. The list of text prompts is generated from labels by simple strings such as ``a photo of $<$ class $>$". In practice, a larger set of text prompts per class is used, we provide details in the supplemental. In our case, we generate the CLIP embeddings and the scores for each of the instance regions $\{M^i_{t}\}$ by using the corresponding box ($\{b^i_{t}\}$) to get the instance crop ($\{{b^i_{t}}^\oplus\}$). We assign initial class labels to each of the instances using the CLIP model as following.
\begin{equation}
     \text{class}(i) = \arg \max_{l\in\mathcal{C}} \left( f^\text{CLIP}_\text{vision}({b^i_{t}}^\oplus) \ \boldsymbol{\cdot}\  f^\text{CLIP}_\text{text}(\text{a photo of  $\langle l \rangle$ })  \right)
\end{equation}
where ${b^i_{t}}^\oplus$ is the cropped instance region for frame  $V_{t}$ and $\text{class}(i)$ is the initial class assigned to the $i^\text{th}$ instance by the CLIP model. We also denote the CLIP class score for this instance as $u^i_{t}=f^\text{CLIP}_\text{vision}({b^i_{t}}^\oplus) \ \boldsymbol{\cdot}\  f^\text{CLIP}_\text{text}(\text{a photo of $ < \text{class}(i) >$)}$. 

\noindent\textbf{Prototype Memory Filtering (PMF).} These initial classes or pseudo-labels are often noisy and contain a lot of false positives.
To address this, we create class specific prototypes using the initial class labels as following. For each class label $l\in\mathcal{C}$ we accumulate all the instance features given by $f^\text{CLIP}_\text{vision}({b^i_{t}}^\oplus)$.
We apply K-Means clustering on the features and compute $k^l$ centroids. We set $k^l$ to be proportional to the number of instances in class $l$. We denote these clusters as the prototype clusters for class $l$. We can now compute an out-of-distribution score for each instance $i$ such that $\text{class}(i)=l$. To do this we compute the cosine similarity between the prototype clusters of class $l$ and CLIP features for an instance $f^\text{CLIP}_\text{vision}({b^i_{t}}^\oplus)$ which has the same predicted class. We discard all instances for which the maximum similarity with any prototype is less than a threshold $\tau$. Together with objectness score and CLIP score, $\tau$ determines the final instances we retain for each prototype in our prototype memory. These protoypes' embeddings collectively reflect various pose, appearances and instances of the objects within the same category.

\subsection{Training the segmentation model}
\label{subsec:mt}
Using the pseudo-mask labels from the last step, we next train an instance segmentation model which comprises of a convolutional image encoder $\mathcal{E}$ and a transformer decoder $\mathcal{D}$. Our setup is similar to the one used in MinVIS~\cite{huang2022minvis} which uses a supervised setting, as compared to our unsupervised case. We provide more details below and some additional details in the supplemental.

Given a frame $V_{t}$ and corresponding pseudo-labels $l^i_{t}$ and $M^i_{t}$, the model uses the fully convolutional image encoder to extract multi-scale features $\mathbf{F}_{t} = \mathcal{E}(V_{t})$.
Input to the decoder are $q \in \mathcal{Q}$ learnable query embeddings along with the the encoder features ($\mathbf{F}_{t}$) with $|\mathcal{Q}|=N_q$.
The transformer decoder then outputs transformed queries such that $\hat{q} = \mathcal{D}(\mathbf{F}_{t}, q)$.
Each query $\hat{q} \in \hat{\mathcal{Q}}$ is passed to a classification head $f_\text{cls}$ to obtain classification scores $s = f_\text{cls}(\hat{q}), s\in \mathbb{R}^{1 \times C}$ where $C=|\mathcal{C}|$.

Along with classification score per query, we also obtain a segmentation masks $M \in \mathbb{R}^{N_q \times H \times W}$ for the query by convolving transformed query embedding $\hat{q}$ with last layer's features in $\mathbf{F}_{t}$ where $H$ and $W$ are the height and width of the image. In other words, $M = \sigma(\hat{q}*\mathbf{F}^{-1}_{t})$ where $\sigma(.)$ is the sigmoid function, $*$ is the convolution operation and $-1$ represents the last layer's features. During training, the classification head outputs ($s$) and segmentation head outputs ($M$) are used to perform bipartite matching between predictions and pseudo-labels that minimize the classification and segmentation losses. Once assigned, the losses are recomputed based on the matching to obtain the total loss $\mathcal{L}_\text{vis} = \mathcal{L}_\text{cls} + \mathcal{L}_\text{seg}$ where $\mathcal{L}_\text{cls}$, the classification loss, is computed using cross entropy. $\mathcal{L}_\text{seg}$, the segmentation loss, is computed using binary cross entropy and dice loss~\cite{milletari2016v}.

Once the model is trained, the per-frame learned queries $\hat{q}$ are used to perform tracking and generate temporally consistent instance masks for each instance along with the predicted class labels.

\subsection{Tracking using learned Queries and Memory}
\label{subsec:tracking}

\noindent\textbf{Query based Tracking.} Our transformer-based model learns queries which help us identify and label each instance region. During training, we utilize per frame predictions to propogate the loss and do not use any temporal cues. But while performing inference, we require temporally and spatially consistent predictions. To extend the per-frame predictions during inference, we utilize the similarity between query embeddings between adjacent frames. Using cosine similarity based Hungarian matching between queries $\mathcal{Q}^t$ and $\mathcal{Q}^{t+1}$ of frames $V_{t}$ and the next frame $V_{t+1}$ we obtain the permutation operator ($P^{t}$) of queries ($\mathcal{Q}^t$) which assigns them to $\mathcal{Q}^{t+1}$.
Utilizing a large number of queries, automatically tackles occlusion, and birth and death of tracklets by detecting null/empty masks, with enough queries remaining to track the foreground objects of interest. The final class prediction in this case for each tracklet is computed using the averaged logits across time.

\noindent\textbf{Tracking with Memory Bank.} While using cosine similarity to propagate instances across frames using Hungarian matching can give us tracklets and corresponding labels, we notice that for the unsupervised case, these are not very accurate. This can be attributed to the noise in pseudo-labels involved in training and the lack of encoding of variations in appearance of the same instance.
To further improve the tracking we utilize a tracking memory module. This module performs averaging of the query vectors based on the matching between two frames to use a weighted query feature from all previous frames. This adds temporal memory to each query feature which is able to encode the instance appearance over a time window instead of just focusing on the previous frame. Specifically, given two frames, $V_{t}$ and $V_{t+1}$ of video $V_n$, having queries $\mathcal{Q}^{t}$ and $\mathcal{Q}^{t+1}$, instead of matching $Q^{t+1}$ with $\mathcal{Q}^{t}$ we define the averaged $\mathcal{Q}_*^{t}$ instead as follows:
\begin{equation}
    \mathcal{Q}_*^{t} = \lambda*P^{t-1}[\mathcal{Q}^{t}] + (1-\lambda)*\mu^{t-1}
\end{equation}
where $\mu^t \in \mathbb{R}^{N_q\times d}$ is the average memory. Here $N_q$ is the number of per frame queries and $d$ is the dimension of each query vector. We define $\mu^{t-1}$ as follows:
\begin{equation}
    \mu^{t-1} = \frac{1}{t-1}\left(\mathcal{Q}^1 + P^1[\mathcal{Q}^2] + P^2[\mathcal{Q}^3] + ... + P^{t-2}[\mathcal{Q}^{t-1}]\right)
\end{equation}


%% file: 4_experiments.tex
\vspace{-0.1in}
\section{Experiments}
\label{sec:experiments}
\vspace{-0.05in}
\begin{table*}[t]
\centering

\footnotesize
\caption{\textbf{Mask ($\mathcal{M}$) / Box ($\mathcal{B}$) / category ($\mathcal{C}$)  vs. our unsupervised setting on validation set of YouTube-VIS 2019~\cite{yang2019video}, YouTube-VIS 2021~\cite{yang2019video}, and OVIS~\cite{qi2021occluded}.} * indicates training in videos without COCO pretrained model weights as initialization using authors' official code. ``I-Sup.'' and ``V-sup.'' indicate the supervision used in the image dataset and the video dataset, respectively. All results below are based on R50 backbone. Our \oursshort achieves decent results in all three datasets without any videos annotations or dense supervision from images.}
\label{tab:main}
\addtolength{\tabcolsep}{-2pt}
\footnotesize
\begin{tabular}{@{}lccccccccccc@{}}
\toprule
Method & Video-Dataset  & COCO  &I-Sup.&V-Sup. & AP   & AP$_{50}$ & AP$_{75}$ & AR$_1$ & AR$_{10}$ \\\midrule
IDOL~\cite{IDOL}            & YTVIS-2019 &\cmark &$\mathcal{M}$&$\mathcal{M}$& 49.5 & 74.0    & 52.9    & 47.7   & 58.7    \\
MinVIS~\cite{huang2022minvis} &YTVIS-2019 &\cmark &$\mathcal{M}$&$\mathcal{M}$ & 47.4 & 69.0    & 52.1    & 45.7   & 55.7    \\
MaskFreeVIS~\cite{ke2023mask}  &YTVIS-2019&\cmark&$\mathcal{B}$ &$\mathcal{B}$ & 42.5 & 66.8    & 45.7    & 41.2   & 51.2    \\
MinVIS~\cite{huang2022minvis}  &YTVIS-2019&\xmark&- &$\mathcal{M}$ & 30.3 & 51.3    & 30.1    & 34.7   & 38.1    \\
\midrule
WISE~\cite{laradji2019masks}     &YTVIS-2019  &\xmark  & -&$\mathcal{C}$   & 6.3  & 17.5     & 3.5     & 7.1    & 7.8     \\
IRN~\cite{ahn2019weakly}     &YTVIS-2019  &\xmark  & - &$\mathcal{C}$   & 7.3  & 18.0     & 3.0     & 9.0    & 10.7     \\
WeakVIS~\cite{liu2021weakly}     &YTVIS-2019   &\xmark &  - & $\mathcal{C}$   & 10.5  & 27.2     & 6.2     & 12.3    & 13.6     \\

DeepSort~\cite{wojke2017simple}     &YTVIS-2019&\xmark &-   & -   & 12.5  & 27.1   & 10.8    & 15.3   & 18.1    \\
\oursshort     &YTVIS-2019 &\xmark   & -  &- & 21.4  & 42.3   & 19.4    & 22.5   & 28.2\\

\midrule
\midrule
IDOL~\cite{IDOL}            & YTVIS-2021 &\cmark &$\mathcal{M}$  & $\mathcal{M}$  & 43.9 & 68.0    & 49.6    & 38.0   & 50.9    \\
MinVIS~\cite{huang2022minvis}   &YTVIS-2021 &\cmark &$\mathcal{M}$      &$\mathcal{M}$& 44.2 & 66.0    & 48.1    & 39.2   & 51.7    \\
MaskFreeVIS~\cite{ke2023mask}  &YTVIS-2021&\cmark&$\mathcal{B}$ &$\mathcal{B}$ & 36.2 & 60.8    & 39.2    & 34.6   & 45.6    \\
MinVIS~\cite{huang2022minvis}    &YTVIS-2021 &\xmark &-      &$\mathcal{M}$& 32.1 & 54.0    & 33.2    & 30.9   & 39.1    \\
\midrule

DeepSort~\cite{wojke2017simple}     &YTVIS-2021 &\xmark &-     & -  & 10.3  & 23.0     & 9.4    & 11.9    & 15.5    \\

\oursshort     &YTVIS-2021   &\xmark    & - &- & 17.5  & 35.6     & 16.3    & 19.7    & 26.3    \\

\midrule
\midrule
IDOL~\cite{IDOL} & OVIS &\cmark   &$\mathcal{M}$     &$\mathcal{M}$  &30.2&	51.3&	30.0&			15.0&		37.5 \\
MinVIS~\cite{huang2022minvis}         & OVIS &\cmark   &$\mathcal{M}$     &$\mathcal{M}$   & 25.0 & 45.5  & 24.0      & 13.9   & 29.7      \\
MaskFreeVIS~\cite{ke2023mask}  &OVIS&\cmark&$\mathcal{M}$ &$\mathcal{B}$ & 15.7 & 35.1    & 13.1    & 10.1   & 20.4    \\
MinVIS~\cite{huang2022minvis}         & OVIS &\xmark  &-      &$\mathcal{M}$   & 15.0 & 33.9  & 12.8      & 9.8   & 19.3      \\
\midrule
DeepSort~\cite{wojke2017simple}              &  OVIS  &\xmark & -&- & 1.6  & 4.0  & 1.4      & 1.9    & 3.9       \\
\oursshort               &  OVIS &\xmark  & -&- & 3.5  & 11.1  & 2.1      & 3.6    & 7.0       \\

\bottomrule
\end{tabular}
\vspace{-0.24in}
\end{table*}


We evaluate our method on three VIS benchmarks: YouTube-VIS 2019~\cite{yang2019video} (YTVIS-2019), YouTube-VIS 2021~\cite{yang2019video}  (YTVIS-2021), and Occluded VIS~\cite{qi2021occluded} (OVIS). We describe our experimental setup in \Cref{sec_exp:setup}, compare \oursshort with state-of-the-art fully-supervised approaches in \Cref{sec_exp:compare}, and provide an ablation study in \Cref{sec_exp:ablation}. For more details, please refer to the supplement.

\vspace{-0.05in}
\subsection{Experimental Setup}
\label{sec_exp:setup}
\vspace{-0.05in}

\noindent\textbf{Datasets.} \textbf{YouTube-VIS 2019} dataset~\cite{yang2019video} (YTVIS-2019) is widely used for video instance segmentation task. It comprises 2,883 labeled videos, 131,000 instance masks, and covers 40 different classes. An improved version called \textbf{YouTube-VIS 2021} (YTVIS-2021) was also introduced~\cite{yang2019video}, featuring 8,171 unique video instances and 232,000 instance masks. \textbf{OVIS} is another challenging dataset, offering heavy occlusion, longer sequences and more number of objects. OVIS consists of 296,000 instance masks and contains an average of 5.8 instances per video across 25 classes.


\noindent\textbf{Experimental Setup.} 
We highlight our experimental setup here. For pseudo-label generation, we utilize CutLER~\cite{wang2023cut} pretrained on ImageNet for class-agnostic masks generation using their Cascade-Mask-RCNN-based pretrained checkpoint. For labeling the proposed regions we use CLIP ViT-bigG-14~\cite{Radford2021LearningTV} from OpenCLIP~\cite{ilharco_gabriel_2021_5143773}. We apply a  threshold of 0.7 to both objectness score from CutLER ($o^i_{t}$) and class score ($u^i_{t}$) from CLIP, and set $\tau=0.7$.

For VIS architecture and optimization, 
we follow MinVIS~\cite{huang2022minvis}'s model architecture, training hyperparameters, and losses. Specifically, for the MinVIS architecture, we utilizes a ResNet-50~\cite{He2015DeepRL} (R50) image encoder and a transformer decoder and sets $N_q=100$. However, we made three major modifications to it. Firstly, instead of relying on ground truth masks, we employed pseudo masks generated by our method (cf. Section~\ref{subsec:pseudo}). Secondly, instead of pretraining on COCO with dense labels, we use ImageNet classification for backbone initialization and train the transformer from scratch. Therefore, we increased the number of interactions to 320k as our setup requires more time to converge. Lastly,we incorporated our proposed tracking memory during inference and set $\lambda$=0.5.

\begin{figure*}[t]
  \centering
  \includegraphics[width=0.9\linewidth]{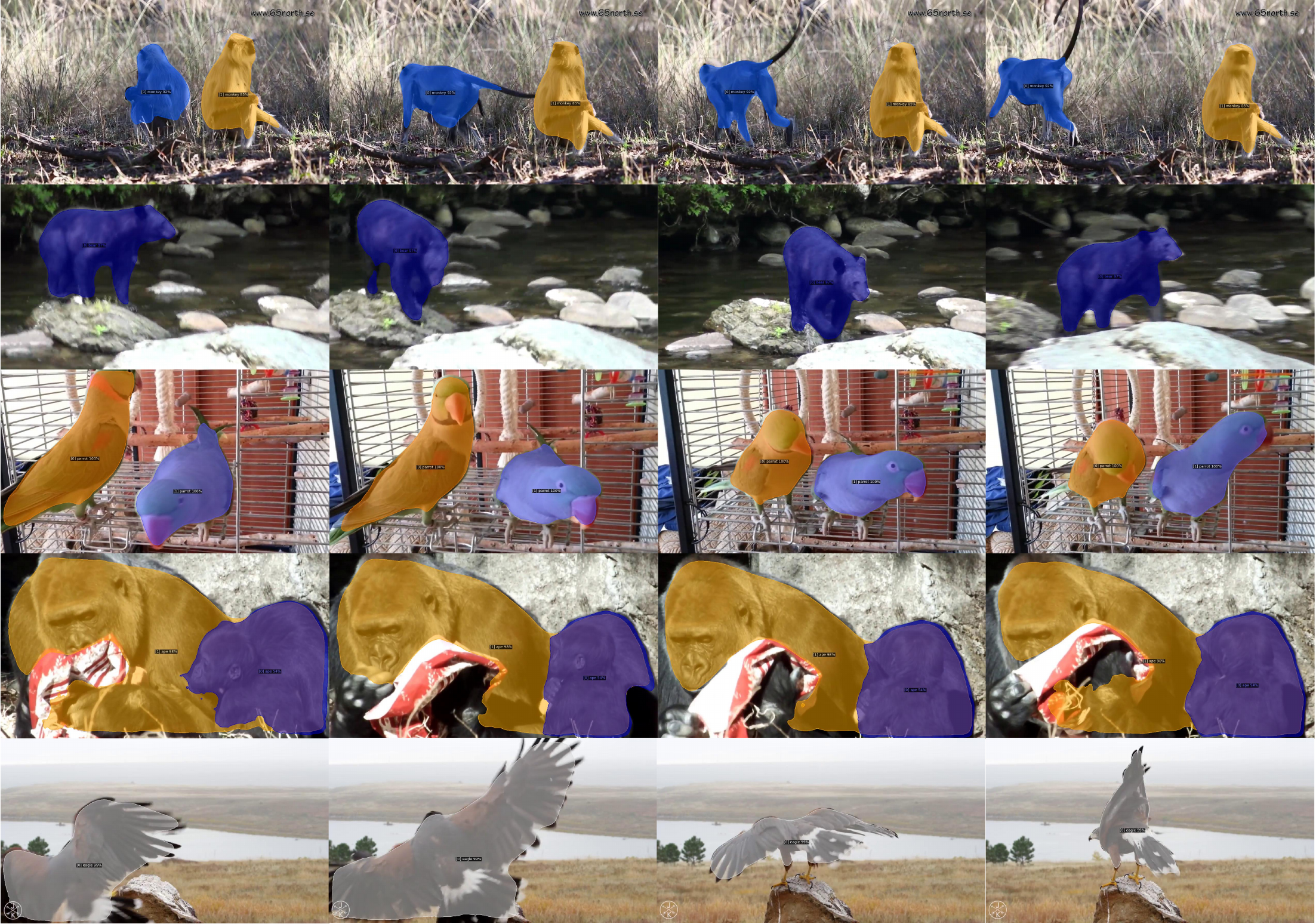}
    \footnotesize
    \caption{\textbf{Visualizations on YoutubeVIS-2019~\cite{yang2019video} with our \oursshort}. Each row shows temporal instance mask and class predictions. Our method is able to work for examples containing multiple instances of the same class (rows 1, 3, 4) and also when there are instances from different classes (row 5). UVIS shows promising results when instances of the same class might overlap (row 4).}
  \label{fig:qualitative}
  \vspace{-0.2in}
\end{figure*}



\noindent\textbf{Baselines.} 
To our best knowledge, ours is one of the first works to introduce the task of unsupervised VIS and does not have any direct baselines to compare with. We propose a new baseline for comparison by utilizing DeepSort~\cite{wojke2017simple}. DeepSort~\cite{wojke2017simple} does not require any training and produces tracks given per-frame detections and deep features. We feed our per-frame pseudo-labels in the validation split and the associated CLIP CLS token features of each instance crop into DeepSort to generate satio-temporal masks for evaluation.


\noindent\textbf{Metrics.} For evaluation, we utilize the metrics of AP (Average Precision) and AR (Average Recall), and evaluate the performance on the validation split in line with the previous work~\cite{IDOL,huang2022minvis,liu2021weakly}.


\vspace{-0.1in}
\subsection{Quantitative Comparison}
\label{sec_exp:compare}
\vspace{-0.1in}

We compare our~\oursshort with recent full-supervised methods including IDOL~\cite{IDOL}, MinVIS~\cite{huang2022minvis} as shown in~\Cref{tab:main}. We also compare our method with recent box-supervised method MaskFreeVIS~\cite{ke2023mask} and category-label supervised method WeakVIS~\cite{liu2021weakly}. Note that MinVIS~\cite{huang2022minvis} (w/o COCO pretraining) serves as the fully-supervised counterpart of \oursshort. 


\noindent\textbf{YouTube-VIS 2019.} 
As shown in Table~\ref{tab:main}, we achieved an impressive AP of $21.4$ without relying on any annotations or COCO pretraining. This result outperforms the previous weakly-supervised method~\cite{liu2021weakly}, which utilized per-frame category labels on videos and external flow networks, by a significant margin of 10.9 AP. Our self-constructed baseline of DeepSort~\cite{wojke2017simple} also performs better than WeakVIS~\cite{liu2021weakly} by 2 AP showing it as an effective approach for comparison.  We also show qualitative results of our approach in~\Cref{fig:qualitative}.

\noindent\textbf{YouTube-VIS 2021.} 
Our \oursshort achieves 17.5 AP on this more challenging dataset. It also beats the DeepSort baseline by 7.2 AP showing the effectiveness of the proposed prototype memory filtering (PMF), training and memory based tracking. These compelling findings highlight the potential of our unsupervised video instance segmentation framework and its ability to deliver competitive results.

\noindent\textbf{Occluded-VIS 2021.} 
In the most challenging setting with the Occluded-VIS 2021 dataset, we achieve a modest result of 3.5 AP despite heavy occlusions and extremely long sequences. This is again a 1.9 AP improvement over the DeepSort baseline.


\begin{table}[t]
\footnotesize
\renewcommand{\arraystretch}{1.2}
\renewcommand{\tabcolsep}{6pt}
\caption{\textbf{Ablation of different components of our pipeline on YTVIS-2019~\cite{yang2019video} val set.}}

	\centering
	\resizebox{1.0\linewidth}{!}{
		\begin{tabular}{@{}ccccccccc@{}}
			\toprule

   			\multirow{2}{*}{Model ID} & \multirow{2}{*}{CLIP} & \multirow{2}{*}{\makecell[cc]{Video\\ Train}} &\multicolumn{3}{c}{Label Denoise} & \multirow{2}{*}{\makecell[cc]{Tracking\\ Memory}} & \multirow{2}{*}{~AP~} & \multirow{2}{*}{~$\Delta$~}  \\  
			\cmidrule{4-6}
			&&&Mask Score& CLIP Score & PMF & \\
			\midrule
			MinVIS~\cite{huang2022minvis} (Upperbound) &- & -  & -& -& -  & - &   30.3   \\
			\midrule
			A1&\cmark\quad &  - & -  & -&- & - &  12.5    & -      \\
			A2&\cmark\quad &  \cmark\quad & -  &-& - & - &  16.6 & 4.1  \\
			A3&\cmark\quad &  \cmark\quad& \cmark\quad  & -&- & - &  18.4 & 5.9  \\
			A4&\cmark\quad &  \cmark\quad& \cmark\quad  & \cmark&- & - &  19.8 & 7.3 \\
			A5&\cmark\quad & \cmark\quad& \cmark\quad  &\cmark&\cmark\quad& - &  20.7 & 8.2 \\
			A6&\cmark\quad & \cmark\quad& \cmark\quad  &\cmark&\cmark\quad& \cmark\quad &  \textbf{21.4} & 8.9  \\
			\bottomrule
		\end{tabular}
	}
	\label{tab:ablation_ind}
 \vspace{-0.2in}
\end{table}

\begin{table*}[t]
\begin{minipage}{.3\textwidth}
\vspace{-1mm}
\renewcommand{\tabcolsep}{8pt}
\renewcommand{\arraystretch}{1.2}
\caption{\textbf{Ablation of Prototype Memory Filtering (PMF) on YouTube-VIS 2019~\cite{yang2019video} val}. Prototype Memory Filtering improves AP by 0.9.}
	\centering
	\resizebox{0.8\linewidth}{!}{
	
        \begin{tabular}{@{}lcccc@{}}
			\toprule                
			${\tau}$ & 0.0 &  0.5 & 0.7 &0.9                    \\
			\midrule
		AP	& 19.8 & 20.1 & \textbf{20.7} & 19.9      \\
			\bottomrule
		\end{tabular}
	}
	\label{tab:ablation_proto}
\end{minipage}
\hfill
\begin{minipage}{.66\textwidth}
\renewcommand{\tabcolsep}{12pt}
\caption{\textbf{Ablation of tracking memory on YouTube-VIS 2019~\cite{yang2019video} val}. Our proposed tracking memory can be generalized to different datasets and different supervision settings.}
	\centering
\vspace{-0.1in}
\resizebox{0.8\textwidth}{!}{
\begin{tabular}{@{}lcccc@{}}
\toprule                
Model &Sup.  &   Dataset   &   AP  &    AP (+Tracking Memory)  \\
\midrule
C0   &-      & YTVIS-2019         & 20.7      & 21.4 \more{(+0.4)}\\
C1   &-  & Occluded VIS    & 3.1      & 3.5 \more{(+0.4)}\\

\midrule
C2   &$\mathcal{I_M}+\mathcal{V_M}$  & YTVIS-2019     & 47.3      & 50.7 \more{(+3.4)} \\
C3   &$\mathcal{I_M}+\mathcal{V_M}$  &Occluded VIS    & 26.7      & 27.2 \more{(+0.5)}  \\

\bottomrule
\end{tabular}
}
 \label{tab:ablation_track}
\end{minipage}
\vspace{-0.12in}
\end{table*}

\begin{figure*}[t]
  \centering
  \includegraphics[width=0.8\linewidth]{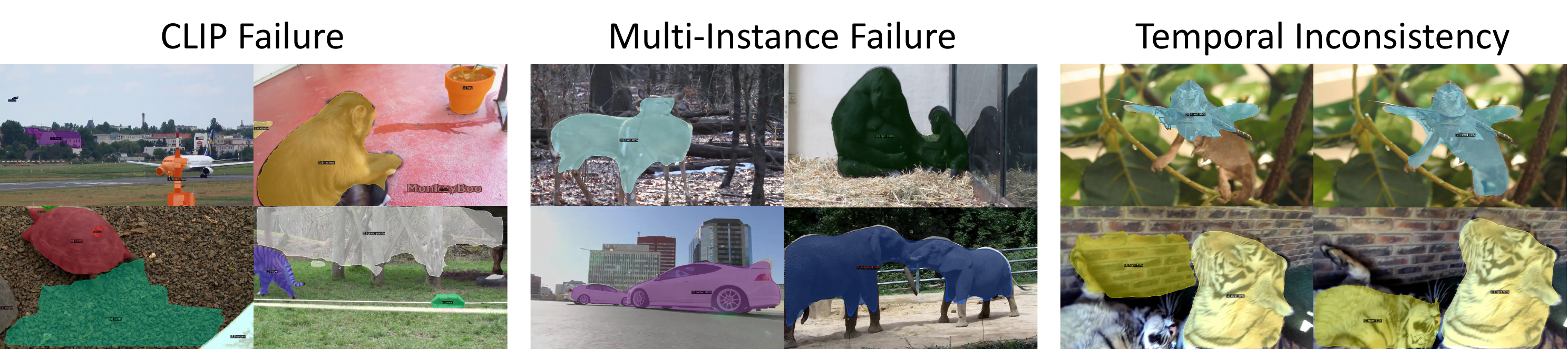}
    \caption{\textbf{Visualizations of failure cases on YoutubeVIS-2019~\cite{yang2019video}}. On the left we show CLIP labeling failures where the CLIP model incorrectly classifies to the wrong class. In the center we show prediction inconsistencies where multiple instances are predicted as one. On the right we show temporal inconsistencies in predicted masks.}
  \label{fig:failure}
  \vspace{-0.2in}
\end{figure*}

\vspace{-0.1in}
\subsection{Ablation Study}
\label{sec_exp:ablation}
\vspace{-0.05in}
We perform ablation on 1) the effects of each model component; 2) the prototype memory filtering design choice; and 3) the tracking memory component generalizability. All results are based on R50 backbone and conducted under the same configuration for a fair comparison.

\noindent\textbf{Effects of model components.} We conducted an ablation study to assess the impact of each component of our model, as presented in Table~\ref{tab:ablation_ind}. The Baseline-DeepSort achieves a validation split performance of 12.5 AP using CLIP features for tracking, without any video training. When training a VIS model with pseudo-labels without any filtering, the performance increases to 16.6 AP. By incorporating mask score and CLIP score for filtering, we observe improvements to 18.4 and 19.8 AP, respectively. Our Prototype Memory Filtering (PMF) component further enhances the performance to 20.7 AP, highlighting the importance of employing prototype memory banks for filtering out noisy labels. Finally, with the addition of our Tracking Memory component, the model achieves a 0.7 AP boost, resulting in a final performance of 21.4 AP without any supervision. This performance is only 8.9 AP lower than the upper bound achieved with full mask supervision in videos.

\noindent\textbf{Prototype memory filtering ablation.} We analyze the prototype memory filtering, shown in Table~\ref{tab:ablation_proto}, by adjusting the threshold for keeping proposals. We observe that a lower threshold is relatively safer and yields improvements (+0.3 AP) compared to not using any prototype memory filtering. As we increase the threshold ($\tau$), the performance further improves (+0.9 AP) due to the removal of noisy labels facilitated by our prototype memory. However, we noticed that an excessively large threshold of 0.9 does not perform as well. This could be attributed to the fact that a higher threshold leads to the significant removal of true positives.

\noindent\textbf{Tracking memory ablation.}
We ablate our tracking memory module in Table~\ref{tab:ablation_track}. In the unsupervised setup, incorporating the tracking memory resulted in a consistent 0.4 AP boost on both the YTVIS-2019 and Occluded VIS datasets. We observe similar consistent improvement in the supervised setup too, where we use our tracking module over the official fully-supervised MinVIS checkpoint and produce a boost of 3.4 AP on YouTube-VIS 2019 and 0.5 AP on Occluded VIS. This result highlights the importance of temporal information compared to MinVIS, which only utilizes information from consecutive frames for tracking. These experimental results confirm the generalization ability and effectiveness of our tracking memory component, both in unsupervised and supervised settings.

\noindent\textbf{Failure cases.} In~\Cref{fig:failure} we highlight some failure examples. We show examples where the CLIP model assigns incorrect class to the region (left). We also show multi-instance failures where the trained model assigns an instance mask covering multiple instances of the same category (center). This usually arises when the two objects occlude each other. Finally, we show temporal inconsistency failures where the model predicts masks that are not temporally consistent and end up not masking the object perfectly.

%% file: 5_conclusion.tex
\vspace{-0.1in}
\section{Conclusion}
\label{sec:conclusion}
\vspace{-0.1in}
We introduced UVIS, the first unsupervised video instance segmentation approach that eliminates the need for video annotations or dense pretraining, to the best of our knowledge. \oursshort consists of three essential steps and incorporates our proposed dual-memory module to improve mask predictions. First, we generate class-agnostic instance masks for each video frame using CutLER and associate them with semantic labels using CLIP. We then employ a class-specific prototype memory bank to filter out noisy labels. Second, we train a transformer-based VIS model using image-based training and pseudo-labels obtained from the previous step. Third, during inference, we connect per-frame predictions to form mask tubes using bipartite matching of query embeddings. We enhance the tracking performance by updating query embeddings using our tracking memory bank, which captures long-term temporal information. We evaluate our approach on three standard benchmarks, namely YTVIS 2019, YTVIS 2021, and OVIS. Our work demonstrates the potential of utilizing foundation models for unsupervised VIS, contributing to the advancement of scalable video applications.

\noindent\textbf{Acknowledgements.} This work was partially funded by
NSF CAREER Award (\#2238769) to AS.

%% file: 0_cvpr24w_ID30.bbl
\begin{thebibliography}{69}
\providecommand{\natexlab}[1]{#1}
\providecommand{\url}[1]{\texttt{#1}}
\expandafter\ifx\csname urlstyle\endcsname\relax
  \providecommand{\doi}[1]{doi: #1}\else
  \providecommand{\doi}{doi: \begingroup \urlstyle{rm}\Url}\fi

\bibitem[Ahn et~al.(2019)Ahn, Cho, and Kwak]{ahn2019weakly}
Jiwoon Ahn, Sunghyun Cho, and Suha Kwak.
\newblock Weakly supervised learning of instance segmentation with inter-pixel
  relations.
\newblock In \emph{Proceedings of the IEEE/CVF conference on computer vision
  and pattern recognition}, pages 2209--2218, 2019.

\bibitem[Amir et~al.(2022)Amir, Gandelsman, Bagon, and Dekel]{amir2021deep}
Shir Amir, Yossi Gandelsman, Shai Bagon, and Tali Dekel.
\newblock Deep vit features as dense visual descriptors.
\newblock \emph{ECCVW What is Motion For?}, 2022.

\bibitem[Biederman(1987)]{biederman1987recognition}
Irving Biederman.
\newblock Recognition-by-components: a theory of human image understanding.
\newblock \emph{Psychological review}, 94\penalty0 (2):\penalty0 115, 1987.

\bibitem[Bommasani et~al.(2021)Bommasani, Hudson, Adeli, Altman, Arora, von
  Arx, Bernstein, Bohg, Bosselut, Brunskill, Brynjolfsson, Buch, Card,
  Castellon, Chatterji, Chen, Creel, Davis, Demszky, Donahue, Doumbouya,
  Durmus, Ermon, Etchemendy, Ethayarajh, Fei-Fei, Finn, Gale, Gillespie, Goel,
  Goodman, Grossman, Guha, Hashimoto, Henderson, Hewitt, Ho, Hong, Hsu, Huang,
  Icard, Jain, Jurafsky, Kalluri, Karamcheti, Keeling, Khani, Khattab, Koh,
  Krass, Krishna, Kuditipudi, Kumar, Ladhak, Lee, Lee, Leskovec, Levent, Li,
  Li, Ma, Malik, Manning, Mirchandani, Mitchell, Munyikwa, Nair, Narayan,
  Narayanan, Newman, Nie, Niebles, Nilforoshan, Nyarko, Ogut, Orr,
  Papadimitriou, Park, Piech, Portelance, Potts, Raghunathan, Reich, Ren, Rong,
  Roohani, Ruiz, Ryan, R'e, Sadigh, Sagawa, Santhanam, Shih, Srinivasan,
  Tamkin, Taori, Thomas, Tram{\`e}r, Wang, Wang, Wu, Wu, Wu, Xie, Yasunaga,
  You, Zaharia, Zhang, Zhang, Zhang, Zhang, Zheng, Zhou, and
  Liang]{Bommasani2021FoundationModels}
Rishi Bommasani, Drew~A. Hudson, Ehsan Adeli, Russ Altman, Simran Arora, Sydney
  von Arx, Michael~S. Bernstein, Jeannette Bohg, Antoine Bosselut, Emma
  Brunskill, Erik Brynjolfsson, S. Buch, Dallas Card, Rodrigo Castellon,
  Niladri~S. Chatterji, Annie~S. Chen, Kathleen~A. Creel, Jared Davis, Dora
  Demszky, Chris Donahue, Moussa Doumbouya, Esin Durmus, Stefano Ermon, John
  Etchemendy, Kawin Ethayarajh, Li Fei-Fei, Chelsea Finn, Trevor Gale,
  Lauren~E. Gillespie, Karan Goel, Noah~D. Goodman, Shelby Grossman, Neel Guha,
  Tatsunori Hashimoto, Peter Henderson, John Hewitt, Daniel~E. Ho, Jenny Hong,
  Kyle Hsu, Jing Huang, Thomas~F. Icard, Saahil Jain, Dan Jurafsky, Pratyusha
  Kalluri, Siddharth Karamcheti, Geoff Keeling, Fereshte Khani, O. Khattab,
  Pang~Wei Koh, Mark~S. Krass, Ranjay Krishna, Rohith Kuditipudi, Ananya Kumar,
  Faisal Ladhak, Mina Lee, Tony Lee, Jure Leskovec, Isabelle Levent, Xiang~Lisa
  Li, Xuechen Li, Tengyu Ma, Ali Malik, Christopher~D. Manning, Suvir~P.
  Mirchandani, Eric Mitchell, Zanele Munyikwa, Suraj Nair, Avanika Narayan,
  Deepak Narayanan, Benjamin Newman, Allen Nie, Juan~Carlos Niebles, Hamed
  Nilforoshan, J.~F. Nyarko, Giray Ogut, Laurel Orr, Isabel Papadimitriou,
  Joon~Sung Park, Chris Piech, Eva Portelance, Christopher Potts, Aditi
  Raghunathan, Robert Reich, Hongyu Ren, Frieda Rong, Yusuf~H. Roohani, Camilo
  Ruiz, Jack Ryan, Christopher R'e, Dorsa Sadigh, Shiori Sagawa, Keshav
  Santhanam, Andy Shih, Krishna~Parasuram Srinivasan, Alex Tamkin, Rohan Taori,
  Armin~W. Thomas, Florian Tram{\`e}r, Rose~E. Wang, William Wang, Bohan Wu,
  Jiajun Wu, Yuhuai Wu, Sang~Michael Xie, Michihiro Yasunaga, Jiaxuan You,
  Matei~A. Zaharia, Michael Zhang, Tianyi Zhang, Xikun Zhang, Yuhui Zhang,
  Lucia Zheng, Kaitlyn Zhou, and Percy Liang.
\newblock On the opportunities and risks of foundation models.
\newblock \emph{ArXiv}, 2021.

\bibitem[Caelles et~al.(2018)Caelles, Montes, Maninis, Chen, Van~Gool, Perazzi,
  and Pont-Tuset]{caelles20182018}
Sergi Caelles, Alberto Montes, Kevis-Kokitsi Maninis, Yuhua Chen, Luc Van~Gool,
  Federico Perazzi, and Jordi Pont-Tuset.
\newblock The 2018 davis challenge on video object segmentation.
\newblock \emph{arXiv preprint arXiv:1803.00557}, 2018.

\bibitem[Caelles et~al.(2019)Caelles, Pont-Tuset, Perazzi, Montes, Maninis, and
  Van~Gool]{caelles20192019}
Sergi Caelles, Jordi Pont-Tuset, Federico Perazzi, Alberto Montes,
  Kevis-Kokitsi Maninis, and Luc Van~Gool.
\newblock The 2019 davis challenge on vos: Unsupervised multi-object
  segmentation.
\newblock \emph{arXiv preprint arXiv:1905.00737}, 2019.

\bibitem[Caron et~al.(2021)Caron, Touvron, Misra, J\'egou, Mairal, Bojanowski,
  and Joulin]{caron2021emerging}
Mathilde Caron, Hugo Touvron, Ishan Misra, Herv\'e J\'egou, Julien Mairal,
  Piotr Bojanowski, and Armand Joulin.
\newblock Emerging properties in self-supervised vision transformers.
\newblock In \emph{Proceedings of the International Conference on Computer
  Vision (ICCV)}, 2021.

\bibitem[Chen et~al.(2020)Chen, Kornblith, Swersky, Norouzi, and
  Hinton]{chen2020big}
Ting Chen, Simon Kornblith, Kevin Swersky, Mohammad Norouzi, and Geoffrey~E
  Hinton.
\newblock Big self-supervised models are strong semi-supervised learners.
\newblock \emph{Advances in neural information processing systems},
  33:\penalty0 22243--22255, 2020.

\bibitem[Chen and He(2021)]{chen2021exploring}
Xinlei Chen and Kaiming He.
\newblock Exploring simple siamese representation learning.
\newblock In \emph{Proceedings of the IEEE/CVF conference on computer vision
  and pattern recognition}, pages 15750--15758, 2021.

\bibitem[Chen et~al.(2021)Chen, Xie, and He]{chen2021empirical}
Xinlei Chen, Saining Xie, and Kaiming He.
\newblock An empirical study of training self-supervised vision transformers.
\newblock In \emph{Proceedings of the IEEE/CVF International Conference on
  Computer Vision}, pages 9640--9649, 2021.

\bibitem[Cheng et~al.(2021)Cheng, Choudhuri, Misra, Kirillov, Girdhar, and
  Schwing]{cheng2021mask2former}
Bowen Cheng, Anwesa Choudhuri, Ishan Misra, Alexander Kirillov, Rohit Girdhar,
  and Alexander~G Schwing.
\newblock Mask2former for video instance segmentation.
\newblock \emph{arXiv preprint arXiv:2112.10764}, 2021.

\bibitem[Cho et~al.(2023)Cho, Lee, Lee, Park, Kim, and Lee]{TMO}
Suhwan Cho, Minhyeok Lee, Seunghoon Lee, Chaewon Park, Donghyeong Kim, and
  Sangyoun Lee.
\newblock Treating motion as option to reduce motion dependency in unsupervised
  video object segmentation.
\newblock In \emph{Proceedings of the IEEE/CVF Winter Conference on
  Applications of Computer Vision}, pages 5140--5149, 2023.

\bibitem[Deng et~al.(2009)Deng, Dong, Socher, Li, Li, and
  Fei-Fei]{deng2009imagenet}
Jia Deng, Wei Dong, Richard Socher, Li-Jia Li, Kai Li, and Li Fei-Fei.
\newblock Imagenet: A large-scale hierarchical image database.
\newblock In \emph{2009 IEEE conference on computer vision and pattern
  recognition}, pages 248--255. Ieee, 2009.

\bibitem[Fu et~al.(2020)Fu, Yang, Liu, Huang, and Shi]{fu2020compfeat}
Yang Fu, Linjie Yang, Ding Liu, Thomas~S Huang, and Humphrey Shi.
\newblock Compfeat: Comprehensive feature aggregation for video instance
  segmentation.
\newblock \emph{arXiv preprint arXiv:2012.03400}, 2020.

\bibitem[Fu et~al.(2021)Fu, Liu, Iqbal, De~Mello, Shi, and
  Kautz]{fu2021learning}
Yang Fu, Sifei Liu, Umar Iqbal, Shalini De~Mello, Humphrey Shi, and Jan Kautz.
\newblock Learning to track instances without video annotations.
\newblock In \emph{CVPR}, 2021.

\bibitem[He et~al.(2023)He, Yang, Wang, Wu, Chen, Huang, Ren, Lim, and
  Shrivastava]{he2023towards}
Bo He, Xitong Yang, Hanyu Wang, Zuxuan Wu, Hao Chen, Shuaiyi Huang, Yixuan Ren,
  Ser-Nam Lim, and Abhinav Shrivastava.
\newblock Towards scalable neural representation for diverse videos.
\newblock In \emph{Proceedings of the IEEE/CVF Conference on Computer Vision
  and Pattern Recognition}, pages 6132--6142, 2023.

\bibitem[He et~al.(2015)He, Zhang, Ren, and Sun]{He2015DeepRL}
Kaiming He, X. Zhang, Shaoqing Ren, and Jian Sun.
\newblock Deep residual learning for image recognition.
\newblock \emph{2016 IEEE Conference on Computer Vision and Pattern Recognition
  (CVPR)}, pages 770--778, 2015.

\bibitem[He et~al.(2022)He, Chen, Xie, Li, Doll{\'a}r, and
  Girshick]{he2022masked}
Kaiming He, Xinlei Chen, Saining Xie, Yanghao Li, Piotr Doll{\'a}r, and Ross
  Girshick.
\newblock Masked autoencoders are scalable vision learners.
\newblock In \emph{Proceedings of the IEEE/CVF Conference on Computer Vision
  and Pattern Recognition}, pages 16000--16009, 2022.

\bibitem[Huang et~al.(2022{\natexlab{a}})Huang, Yu, and
  Anandkumar]{huang2022minvis}
De-An Huang, Zhiding Yu, and Anima Anandkumar.
\newblock Minvis: A minimal video instance segmentation framework without
  video-based training.
\newblock \emph{arXiv preprint arXiv:2208.02245}, 2022{\natexlab{a}}.

\bibitem[Huang et~al.(2019)Huang, Wang, Zhang, Yan, and He]{huang2019dynamic}
Shuaiyi Huang, Qiuyue Wang, Songyang Zhang, Shipeng Yan, and Xuming He.
\newblock Dynamic context correspondence network for semantic alignment.
\newblock In \emph{Proceedings of the IEEE/CVF International Conference on
  Computer Vision}, pages 2010--2019, 2019.

\bibitem[Huang et~al.(2022{\natexlab{b}})Huang, Yang, He, Zhang, He, and
  Shrivastava]{huang2022learning}
Shuaiyi Huang, Luyu Yang, Bo He, Songyang Zhang, Xuming He, and Abhinav
  Shrivastava.
\newblock Learning semantic correspondence with sparse annotations.
\newblock In \emph{European Conference on Computer Vision}, pages 267--284.
  Springer, 2022{\natexlab{b}}.

\bibitem[Huang et~al.(2024)Huang, Huang, Yu, Lan, Radhakrishnan, Alvarez,
  Shrivastava, and Anandkumar]{huang2024point}
Shuaiyi Huang, De-An Huang, Zhiding Yu, Shiyi Lan, Subhashree Radhakrishnan,
  Jose~M Alvarez, Abhinav Shrivastava, and Anima Anandkumar.
\newblock What is point supervision worth in video instance segmentation?
\newblock \emph{arXiv preprint arXiv:2404.01990}, 2024.

\bibitem[Hwang et~al.(2021)Hwang, Heo, Oh, and Kim]{hwang2021video}
Sukjun Hwang, Miran Heo, Seoung~Wug Oh, and Seon~Joo Kim.
\newblock Video instance segmentation using inter-frame communication
  transformers.
\newblock \emph{NeurIPS}, 2021.

\bibitem[Ilharco et~al.(2021)Ilharco, Wortsman, Wightman, Gordon, Carlini,
  Taori, Dave, Shankar, Namkoong, Miller, Hajishirzi, Farhadi, and
  Schmidt]{ilharco_gabriel_2021_5143773}
Gabriel Ilharco, Mitchell Wortsman, Ross Wightman, Cade Gordon, Nicholas
  Carlini, Rohan Taori, Achal Dave, Vaishaal Shankar, Hongseok Namkoong, John
  Miller, Hannaneh Hajishirzi, Ali Farhadi, and Ludwig Schmidt.
\newblock Openclip, 2021.
\newblock If you use this software, please cite it as below.

\bibitem[Ji et~al.(2024)Ji, Liang, Zeng, Luo, Xu, Guo, Zheng, Huang, Sun, and
  Xu]{ji2024ace}
Tianying Ji, Yongyuan Liang, Yan Zeng, Yu Luo, Guowei Xu, Jiawei Guo, Ruijie
  Zheng, Furong Huang, Fuchun Sun, and Huazhe Xu.
\newblock Ace : Off-policy actor-critic with causality-aware entropy
  regularization, 2024.

\bibitem[Ke et~al.(2023)Ke, Danelljan, Ding, Tai, Tang, and Yu]{ke2023mask}
Lei Ke, Martin Danelljan, Henghui Ding, Yu-Wing Tai, Chi-Keung Tang, and Fisher
  Yu.
\newblock Mask-free video instance segmentation.
\newblock \emph{arXiv preprint arXiv:2303.15904}, 2023.

\bibitem[Laradji et~al.(2019)Laradji, Vazquez, and Schmidt]{laradji2019masks}
Issam~H Laradji, David Vazquez, and Mark Schmidt.
\newblock Where are the masks: Instance segmentation with image-level
  supervision.
\newblock \emph{arXiv preprint arXiv:1907.01430}, 2019.

\bibitem[Lee et~al.(2022)Lee, Cho, Lee, Park, and Lee]{Lee2022UnsupervisedVO}
Minhyeok Lee, Suhwan Cho, Seung-Hyun Lee, Chaewon Park, and Sangyoun Lee.
\newblock Unsupervised video object segmentation via prototype memory network.
\newblock \emph{2023 IEEE/CVF Winter Conference on Applications of Computer
  Vision (WACV)}, pages 5913--5923, 2022.

\bibitem[Li et~al.(2021)Li, Li, Li, and Zhang]{li2021spatial}
Minghan Li, Shuai Li, Lida Li, and Lei Zhang.
\newblock Spatial feature calibration and temporal fusion for effective
  one-stage video instance segmentation.
\newblock In \emph{Proceedings of the IEEE/CVF Conference on Computer Vision
  and Pattern Recognition}, pages 11215--11224, 2021.

\bibitem[Liu et~al.(2021)Liu, Ramanathan, Mahajan, Yuille, and
  Yang]{liu2021weakly}
Qing Liu, Vignesh Ramanathan, Dhruv Mahajan, Alan Yuille, and Zhenheng Yang.
\newblock Weakly supervised instance segmentation for videos with temporal mask
  consistency.
\newblock In \emph{Proceedings of the IEEE/CVF Conference on Computer Vision
  and Pattern Recognition}, pages 13968--13978, 2021.

\bibitem[L\"uddecke and Ecker(2022)]{lueddecke22_cvpr}
Timo L\"uddecke and Alexander Ecker.
\newblock Image segmentation using text and image prompts.
\newblock In \emph{Proceedings of the IEEE/CVF Conference on Computer Vision
  and Pattern Recognition (CVPR)}, pages 7086--7096, 2022.

\bibitem[Luo et~al.(2022)Luo, Bao, Wu, He, and Li]{Luo2022SegCLIPPA}
Huaishao Luo, Junwei Bao, Youzheng Wu, Xiaodong He, and Tianrui Li.
\newblock Segclip: Patch aggregation with learnable centers for open-vocabulary
  semantic segmentation.
\newblock \emph{ArXiv}, abs/2211.14813, 2022.

\bibitem[Maag et~al.(2021)Maag, Rottmann, Varghese, H{\"u}ger, Schlicht, and
  Gottschalk]{maag2021improving}
Kira Maag, Matthias Rottmann, Serin Varghese, Fabian H{\"u}ger, Peter Schlicht,
  and Hanno Gottschalk.
\newblock Improving video instance segmentation by light-weight temporal
  uncertainty estimates.
\newblock In \emph{2021 International Joint Conference on Neural Networks
  (IJCNN)}, pages 1--8. IEEE, 2021.

\bibitem[Milletari et~al.(2016)Milletari, Navab, and Ahmadi]{milletari2016v}
Fausto Milletari, Nassir Navab, and Seyed-Ahmad Ahmadi.
\newblock V-net: Fully convolutional neural networks for volumetric medical
  image segmentation.
\newblock In \emph{3DV}, 2016.

\bibitem[Perazzi et~al.(2016)Perazzi, Pont-Tuset, McWilliams, {Van Gool},
  Gross, and Sorkine-Hornung]{Perazzi2016}
F. Perazzi, J. Pont-Tuset, B. McWilliams, L. {Van Gool}, M. Gross, and A.
  Sorkine-Hornung.
\newblock A benchmark dataset and evaluation methodology for video object
  segmentation.
\newblock In \emph{Computer Vision and Pattern Recognition}, 2016.

\bibitem[Pont-Tuset et~al.(2017)Pont-Tuset, Perazzi, Caelles, Arbel\'aez,
  Sorkine-Hornung, and {Van Gool}]{Pont-Tuset_arXiv_2017}
Jordi Pont-Tuset, Federico Perazzi, Sergi Caelles, Pablo Arbel\'aez, Alexander
  Sorkine-Hornung, and Luc {Van Gool}.
\newblock The 2017 davis challenge on video object segmentation.
\newblock \emph{arXiv:1704.00675}, 2017.

\bibitem[Qi et~al.(2021)Qi, Gao, Hu, Wang, Liu, Bai, Belongie, Yuille, Torr,
  and Bai]{qi2021occluded}
Jiyang Qi, Yan Gao, Yao Hu, Xinggang Wang, Xiaoyu Liu, Xiang Bai, Serge
  Belongie, Alan Yuille, Philip Torr, and Song Bai.
\newblock Occluded video instance segmentation: A benchmark.
\newblock \emph{arXiv preprint arXiv:2102.01558}, 2021.

\bibitem[Radford et~al.(2021{\natexlab{a}})Radford, Kim, Hallacy, Ramesh, Goh,
  Agarwal, Sastry, Askell, Mishkin, Clark, Krueger, and
  Sutskever]{Radford2021LearningTV}
Alec Radford, Jong~Wook Kim, Chris Hallacy, Aditya Ramesh, Gabriel Goh,
  Sandhini Agarwal, Girish Sastry, Amanda Askell, Pamela Mishkin, Jack Clark,
  Gretchen Krueger, and Ilya Sutskever.
\newblock Learning transferable visual models from natural language
  supervision.
\newblock In \emph{International Conference on Machine Learning},
  2021{\natexlab{a}}.

\bibitem[Radford et~al.(2021{\natexlab{b}})Radford, Kim, Hallacy, Ramesh, Goh,
  Agarwal, Sastry, Askell, Mishkin, Clark, et~al.]{radford2021learning}
Alec Radford, Jong~Wook Kim, Chris Hallacy, Aditya Ramesh, Gabriel Goh,
  Sandhini Agarwal, Girish Sastry, Amanda Askell, Pamela Mishkin, Jack Clark,
  et~al.
\newblock Learning transferable visual models from natural language
  supervision.
\newblock In \emph{International conference on machine learning}, pages
  8748--8763. PMLR, 2021{\natexlab{b}}.

\bibitem[Rambhatla et~al.(2023)Rambhatla, Misra, Chellappa, and
  Shrivastava]{rambhatla2023most}
Sai~Saketh Rambhatla, Ishan Misra, Rama Chellappa, and Abhinav Shrivastava.
\newblock Most: Multiple object localization with self-supervised transformers
  for object discovery, 2023.

\bibitem[Shah et~al.(2020)Shah, Eysenbach, Kahn, Rhinehart, and
  Levine]{shah2020ving}
Dhruv Shah, Benjamin Eysenbach, Gregory Kahn, Nicholas Rhinehart, and Sergey
  Levine.
\newblock Ving: Learning open-world navigation with visual goals.
\newblock \emph{arXiv preprint arXiv:2012.09812}, 2020.

\bibitem[Shin et~al.(2023)Shin, Xie, and Albanie]{shin2022namedmask}
Gyungin Shin, Weidi Xie, and Samuel Albanie.
\newblock Namedmask: Distilling segmenters from complementary foundation
  models.
\newblock In \emph{CVPRW}, 2023.

\bibitem[Sim\'eoni et~al.(2021)Sim\'eoni, Puy, Vo, Roburin, Gidaris, Bursuc,
  P\'erez, Marlet, and Ponce]{LOST}
Oriane Sim\'eoni, Gilles Puy, Huy~V. Vo, Simon Roburin, Spyros Gidaris, Andrei
  Bursuc, Patrick P\'erez, Renaud Marlet, and Jean Ponce.
\newblock Localizing objects with self-supervised transformers and no labels.
\newblock 2021.

\bibitem[Sucheng et~al.(2021)Sucheng, Wenxi, Yongtuo, Haoxin, Guoqiang, and
  Shengfeng]{ren2020rtnet}
Ren Sucheng, Liu Wenxi, Liu Yongtuo, Chen Haoxin, Han Guoqiang, and He
  Shengfeng.
\newblock Reciprocal transformations for unsupervised video object
  segmentation.
\newblock In \emph{Proceedings of the IEEE/CVF Conference on Computer Vision
  and Pattern Recognition}, 2021.

\bibitem[Sun et~al.(2022)Sun, Zheng, Wang, Cohen, and Huang]{sun2022transfer}
Yanchao Sun, Ruijie Zheng, Xiyao Wang, Andrew Cohen, and Furong Huang.
\newblock Transfer rl across observation feature spaces via model-based
  regularization, 2022.

\bibitem[Wang et~al.(2023)Wang, Girdhar, Yu, and Misra]{wang2023cut}
Xudong Wang, Rohit Girdhar, Stella~X Yu, and Ishan Misra.
\newblock Cut and learn for unsupervised object detection and instance
  segmentation.
\newblock \emph{arXiv preprint arXiv:2301.11320}, 2023.

\bibitem[Wang et~al.(2024)Wang, Zheng, Sun, Jia, Wongkamjan, Xu, and
  Huang]{wang2023coplanner}
Xiyao Wang, Ruijie Zheng, Yanchao Sun, Ruonan Jia, Wichayaporn Wongkamjan,
  Huazhe Xu, and Furong Huang.
\newblock {COP}lanner: Plan to roll out conservatively but to explore
  optimistically for model-based {RL}.
\newblock In \emph{The Twelfth International Conference on Learning
  Representations}, 2024.

\bibitem[Wang et~al.(2018)Wang, Hu, Liew, and Wang]{8650146}
Yufei Wang, Yongjiang Hu, Alan Wee-Chung Liew, and Junhu Wang.
\newblock Weakly supervised video object segmentation.
\newblock In \emph{TENCON 2018 - 2018 IEEE Region 10 Conference}, pages
  0315--0320, 2018.

\bibitem[Wang et~al.(2021)Wang, Xu, Wang, Shen, Cheng, Shen, and
  Xia]{wang2021end}
Yuqing Wang, Zhaoliang Xu, Xinlong Wang, Chunhua Shen, Baoshan Cheng, Hao Shen,
  and Huaxia Xia.
\newblock End-to-end video instance segmentation with transformers.
\newblock In \emph{CVPR}, 2021.

\bibitem[Wang et~al.(2022)Wang, Shen, Hu, Yuan, Crowley, and
  Vaufreydaz]{wang2022tokencut}
Yangtao Wang, Xi Shen, Shell~Xu Hu, Yuan Yuan, James~L. Crowley, and Dominique
  Vaufreydaz.
\newblock Self-supervised transformers for unsupervised object discovery using
  normalized cut.
\newblock In \emph{Conference on Computer Vision and Pattern Recognition},
  2022.

\bibitem[Wei et~al.(2022)Wei, Lang, Liang, Feng, Wang, and
  Chen]{10.1145/3506716}
Lili Wei, Congyan Lang, Liqian Liang, Songhe Feng, Tao Wang, and Shidi Chen.
\newblock Weakly supervised video object segmentation via dual-attention
  cross-branch fusion.
\newblock \emph{ACM Trans. Intell. Syst. Technol.}, 13\penalty0 (3), 2022.

\bibitem[Wei et~al.(2023)Wei, Sun, Zheng, Vemprala, Bonatti, Chen, Madaan, Ba,
  Kapoor, and Ma]{wei2023imitation}
Yao Wei, Yanchao Sun, Ruijie Zheng, Sai Vemprala, Rogerio Bonatti, Shuhang
  Chen, Ratnesh Madaan, Zhongjie Ba, Ashish Kapoor, and Shuang Ma.
\newblock Is imitation all you need? generalized decision-making with
  dual-phase training.
\newblock In \emph{Proceedings of the IEEE/CVF International Conference on
  Computer Vision (ICCV)}, pages 16221--16231, 2023.

\bibitem[Wojke et~al.(2017)Wojke, Bewley, and Paulus]{wojke2017simple}
Nicolai Wojke, Alex Bewley, and Dietrich Paulus.
\newblock Simple online and realtime tracking with a deep association metric.
\newblock In \emph{2017 IEEE international conference on image processing
  (ICIP)}, pages 3645--3649. IEEE, 2017.

\bibitem[Wu et~al.(2022)Wu, Liu, Jiang, Bai, Yuille, and Bai]{IDOL}
Junfeng Wu, Qihao Liu, Yi Jiang, Song Bai, Alan Yuille, and Xiang Bai.
\newblock In defense of online models for video instance segmentation.
\newblock In \emph{ECCV}, 2022.

\bibitem[Xu et~al.(2024)Xu, Zheng, Liang, Wang, Yuan, Ji, Luo, Liu, Yuan, Hua,
  Li, Ze, III, Huang, and Xu]{xu2024drm}
Guowei Xu, Ruijie Zheng, Yongyuan Liang, Xiyao Wang, Zhecheng Yuan, Tianying
  Ji, Yu Luo, Xiaoyu Liu, Jiaxin Yuan, Pu Hua, Shuzhen Li, Yanjie Ze,
  Hal~Daum{\'e} III, Furong Huang, and Huazhe Xu.
\newblock Drm: Mastering visual reinforcement learning through dormant ratio
  minimization.
\newblock In \emph{The Twelfth International Conference on Learning
  Representations}, 2024.

\bibitem[Xu et~al.(2018)Xu, Yang, Fan, Yue, Liang, Yang, and
  Huang]{xu2018youtube}
Ning Xu, Linjie Yang, Yuchen Fan, Dingcheng Yue, Yuchen Liang, Jianchao Yang,
  and Thomas Huang.
\newblock Youtube-vos: A large-scale video object segmentation benchmark.
\newblock \emph{arXiv preprint arXiv:1809.03327}, 2018.

\bibitem[Yang et~al.(2019)Yang, Fan, and Xu]{yang2019video}
Linjie Yang, Yuchen Fan, and Ning Xu.
\newblock Video instance segmentation.
\newblock In \emph{Proceedings of the IEEE/CVF International Conference on
  Computer Vision}, pages 5188--5197, 2019.

\bibitem[Yang et~al.(2021{\natexlab{a}})Yang, Fang, Wang, Li, Fang, Shan, Feng,
  and Liu]{yang2021crossover}
Shusheng Yang, Yuxin Fang, Xinggang Wang, Yu Li, Chen Fang, Ying Shan, Bin
  Feng, and Wenyu Liu.
\newblock Crossover learning for fast online video instance segmentation.
\newblock In \emph{ICCV}, 2021{\natexlab{a}}.

\bibitem[Yang et~al.(2021{\natexlab{b}})Yang, Fang, Wang, Li, Shan, Feng, and
  Liu]{yang2021tracking}
Shusheng Yang, Yuxin Fang, Xinggang Wang, Yu Li, Ying Shan, Bin Feng, and Wenyu
  Liu.
\newblock Tracking instances as queries.
\newblock \emph{arXiv preprint arXiv:2106.11963}, 2021{\natexlab{b}}.

\bibitem[Zbontar et~al.(2021)Zbontar, Jing, Misra, LeCun, and
  Deny]{zbontar2021barlow}
Jure Zbontar, Li Jing, Ishan Misra, Yann LeCun, and St{\'e}phane Deny.
\newblock Barlow twins: Self-supervised learning via redundancy reduction.
\newblock In \emph{International Conference on Machine Learning}, pages
  12310--12320. PMLR, 2021.

\bibitem[Zhen et~al.(2020)Zhen, Li, Zhou, Shang, Feng, Fang, and
  Quan]{Zhen2020LearningDF}
Mingmin Zhen, Shiwei Li, Lei Zhou, Jiaxiang Shang, Haoan Feng, Tian Fang, and
  Long Quan.
\newblock Learning discriminative feature with crf for unsupervised video
  object segmentation.
\newblock In \emph{European Conference on Computer Vision}, 2020.

\bibitem[Zheng et~al.(2023)Zheng, Wang, Xu, and Huang]{zheng2023is}
Ruijie Zheng, Xiyao Wang, Huazhe Xu, and Furong Huang.
\newblock Is model ensemble necessary? model-based {RL} via a single model with
  lipschitz regularized value function.
\newblock In \emph{The Eleventh International Conference on Learning
  Representations}, 2023.

\bibitem[Zheng et~al.(2024{\natexlab{a}})Zheng, Cheng, III, Huang, and
  Kolobov]{zheng2024prise}
Ruijie Zheng, Ching-An Cheng, Hal~Daumé III, Furong Huang, and Andrey Kolobov.
\newblock Prise: Learning temporal action abstractions as a sequence
  compression problem, 2024{\natexlab{a}}.

\bibitem[Zheng et~al.(2024{\natexlab{b}})Zheng, Liang, Wang, Ma, Daum{\'e}~III,
  Xu, Langford, Palanisamy, Basu, and Huang]{zheng2024premier}
Ruijie Zheng, Yongyuan Liang, Xiyao Wang, Shuang Ma, Hal Daum{\'e}~III, Huazhe
  Xu, John Langford, Praveen Palanisamy, Kalyan~Shankar Basu, and Furong Huang.
\newblock Premier-taco: Pretraining multitask representation via temporal
  action-driven contrastive loss.
\newblock \emph{arXiv preprint arXiv:2402.06187}, 2024{\natexlab{b}}.

\bibitem[Zheng et~al.(2024{\natexlab{c}})Zheng, Wang, Sun, Ma, Zhao, Xu,
  Daum{\'e}~III, and Huang]{zheng2024texttt}
Ruijie Zheng, Xiyao Wang, Yanchao Sun, Shuang Ma, Jieyu Zhao, Huazhe Xu, Hal
  Daum{\'e}~III, and Furong Huang.
\newblock Taco: Temporal latent action-driven contrastive loss for visual
  reinforcement learning.
\newblock \emph{Advances in Neural Information Processing Systems}, 36,
  2024{\natexlab{c}}.

\bibitem[Zhou et~al.(2021{\natexlab{a}})Zhou, Loy, and Dai]{Zhou2021ExtractFD}
Chong Zhou, Chen~Change Loy, and Bo Dai.
\newblock Extract free dense labels from clip.
\newblock In \emph{European Conference on Computer Vision}, 2021{\natexlab{a}}.

\bibitem[Zhou et~al.(2021{\natexlab{b}})Zhou, Li, Li, and Shao]{zhou2021target}
Tianfei Zhou, Jianwu Li, Xueyi Li, and Ling Shao.
\newblock Target-aware object discovery and association for unsupervised video
  multi-object segmentation.
\newblock In \emph{Proceedings of the IEEE/CVF Conference on Computer Vision
  and Pattern Recognition}, pages 6985--6994, 2021{\natexlab{b}}.

\bibitem[Zhou et~al.(2023)Zhou, Lei, Zhang, Liu, and Liu]{zhou2022zegclip}
Ziqin Zhou, Yinjie Lei, Bowen Zhang, Lingqiao Liu, and Yifan Liu.
\newblock Zegclip: Towards adapting clip for zero-shot semantic segmentation.
\newblock \emph{Proceedings of the IEEE/CVF Conference on Computer Vision and
  Pattern Recognition (CVPR)}, 2023.

\bibitem[Zhu et~al.(2021)Zhu, Zhu, Liang, and Chang]{zhu2021deep}
Fengda Zhu, Yi Zhu, Xiaodan Liang, and Xiaojun Chang.
\newblock Deep learning for embodied vision navigation: A survey.
\newblock \emph{arXiv preprint arXiv:2108.04097}, 2021.

\end{thebibliography}
